\pdfoutput=1

\documentclass[11pt]{article}

\usepackage[]{acl2023}

\usepackage{times}
\usepackage{latexsym}
\usepackage{subfiles}
\usepackage{graphicx}
\usepackage{booktabs}
\usepackage{diagbox}
\usepackage{slashbox}
\usepackage{float}
\usepackage{textcomp}
\usepackage{enumitem}
\usepackage{xcolor}
\usepackage{colortbl}
\usepackage{todonotes}
\usepackage{multirow}
\usepackage{subcaption}
\usepackage{amsmath}
\usepackage{dsfont}
\usepackage{algorithm}
\usepackage{algorithmic}
\usepackage{algorithmic}
\usepackage{amssymb}
\usepackage[capitalise]{cleveref}

\usepackage[T1]{fontenc}

\usepackage[utf8]{inputenc}

\usepackage{microtype}

\newcommand{\Skip}[1]{}

\newcommand{\ie}{\textit{i}.\textit{e}.\ }
\newcommand{\eg}{\textit{e}.\textit{g}.\ }

\newcommand{\numdata}{650}

\newcommand{\secref}[1]{Section \ref{#1}}
\newcommand{\figref}[1]{Figure \ref{#1}}
\newcommand{\tbref}[1]{Table \ref{#1}}

\newcommand{\appendixref}[1]{Append. Sec. \ref{#1}}

\newcommand{\dotieconcat}[2]{
  \text{\raisebox{.8ex}{$\smallfrown$}}%
}

\newcommand{\mypar}[1]{\noindent\textbf{#1}}

\title{Learning Action Conditions from Instructional Manuals \\ for Instruction Understanding}

{\centering
    \author{
        Te-Lin Wu$^{1}$,
        Caiqi Zhang$^2$,
        Qingyuan Hu$^{1}$,
        Alex Spangher$^{3}$,
        Nanyun Peng$^1$
        \\
    $^1$University of California, Los Angeles,
    $^2$University of Cambridge, \\
    $^3$Information Sciences Institute, University of Southern California\\
    \texttt{\{telinwu,violetpeng,hu528\}@cs.ucla.edu}, \\ \texttt{cz391@cam.ac.uk},
    \texttt{spangher@isi.edu} \\
}
}

\begin{document}

\maketitle

\begin{abstract}
    The ability to infer pre- and postconditions of an action is vital for comprehending complex instructions, and is essential for applications such as autonomous instruction-guided agents and assistive AI that supports humans to perform physical tasks.
In this work, we propose a task dubbed action condition inference, which extracts mentions of preconditions and postconditions of actions in instructional manuals. 
We propose a weakly supervised approach utilizing automatically constructed large-scale training instances from online instructions, and curate a densely human-annotated and validated dataset to study how well the current NLP models do on the proposed task. 
We design two types of models differ by whether contextualized and global information is leveraged, as well as various combinations of heuristics to construct the weak supervisions.
Our experiments show a \textgreater 20\% F1-score improvement with considering the entire instruction contexts and a \textgreater~6\% F1-score benefit with the proposed heuristics. However, the best performing model is still well-behind human performance.\footnote{The curated resource can be downloaded at: \href{https://github.com/telin0411/action-conditions.git}{here}.}
\end{abstract}

\section{Introduction}

\begin{figure}[t!]
\centering
    \includegraphics[width=1.0\columnwidth]{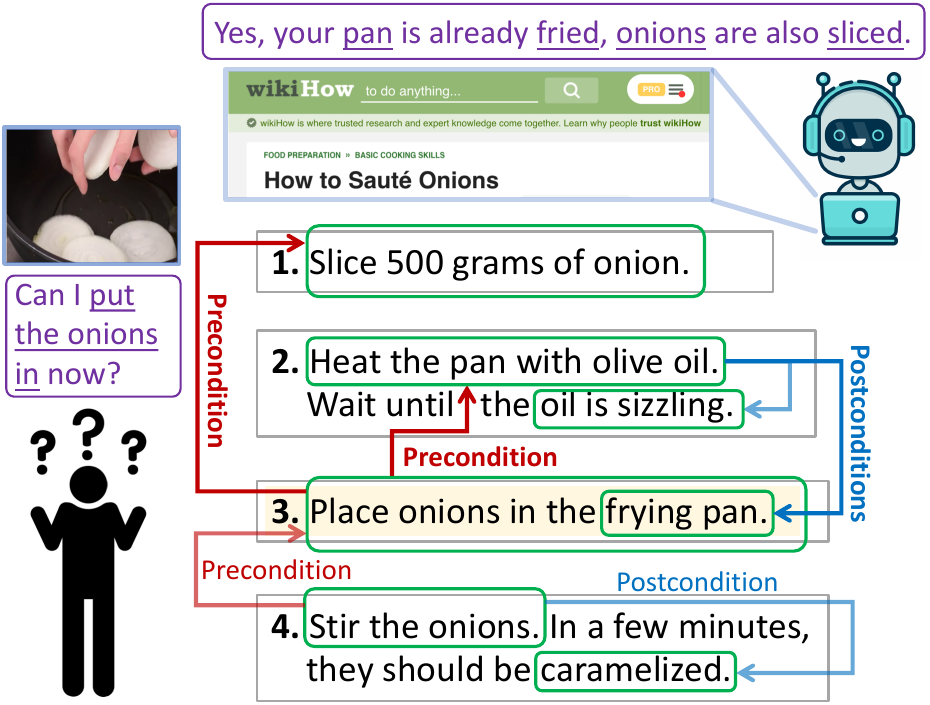}
    \caption{
    \footnotesize
    \textbf{The Action Condition Inference Task:}
    We propose a task that probes models' ability to infer both \textit{preconditions} and \textit{postconditions} of an \textit{action} from instructional manuals. It has wide applications to \eg assistive AI and task-solving robots.
    $^{*}$This instruction is simplified for illustration.
    }
    \label{fig:teaser}
\end{figure}

When performing complex tasks (\eg \textit{making a gourmet dish}), instructional manuals are often referred to as useful guidelines. 
To follow the instructed actions, it is crucial to 
understand the \textit{preconditions}, \ie prerequisites before taking a particular action, and the \textit{postconditions}, \ie the status supposed to be reached after performing the action.
Knowledge of action-condition dependencies is prevalent and inferable in many instructional texts.
For example, in~\figref{fig:teaser}, before performing the action ``\textit{place onions}" in step 3, both \textit{preconditions}: ``\textit{heat the pan}" (in step 2) and ``\textit{slice onions}" (in step 1) have to be successfully accomplished.
Likewise, executing ``\textit{stir onions}" (in step 4), leads to its \textit{postcondition}, ``\textit{caramelized}" (also in step 4).

\definecolor{LightBlue}{rgb}{0.78,1,1}
\definecolor{LightYellow}{rgb}{1,1,0.68}
\definecolor{LightGreen}{rgb}{0.78,1,0.78}

\begin{figure*}[t!]
\centering
    \includegraphics[width=.99\textwidth]{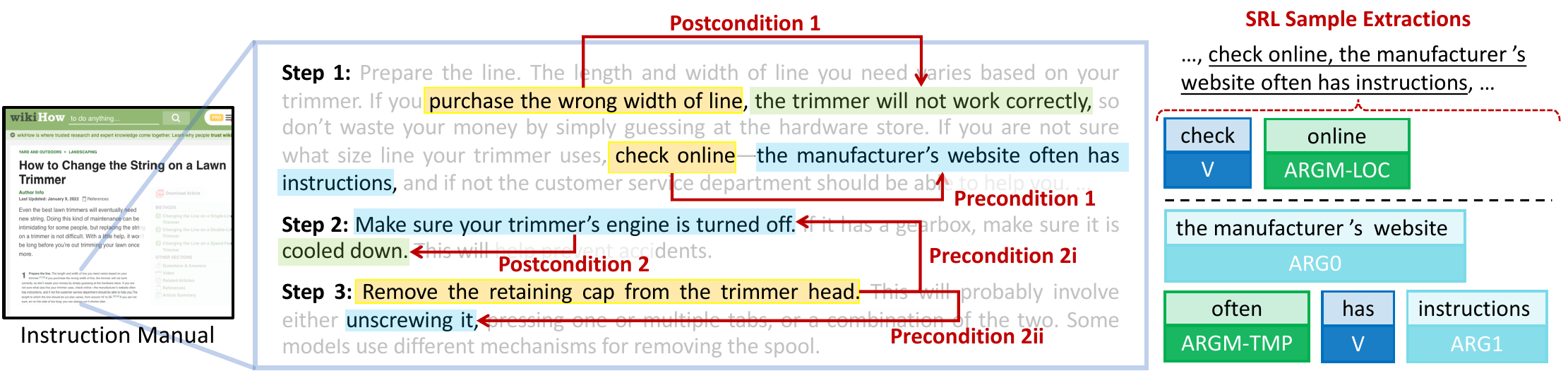}
    \vspace{-.3em}
    \caption{
    \footnotesize
    \textbf{Terminologies:}
    \textbf{(Left)} shows a few exemplar\colorbox{LightYellow}{actionables}with their associated\colorbox{LightBlue}{preconditions}and\colorbox{LightGreen}{postconditions}.
    Notice that an actionable can have multiple pre- or postconditions and they can span across different instruction steps
    (for simplicity we do not show an exhausted set of text segments, and the actual instruction contexts are much longer).
    \textbf{(Right)} SRL is used to postulate the text segments (actionables and conditions).
    We show a sample SRL extraction corresponding to one of the dependency linkages on the left.
    The SRL~\texttt{ARG} labels also provide useful information for designing our heuristics (\secref{sec:data_heus}).
    }
    \label{fig:term}
\end{figure*}

For autonomous agents or assistant AI that aids humans to accomplish tasks, understanding the conditions provides a structured view of a task~\cite{linden1994generating, aeronautiques1998pddl, branavan2012learning, sharma2020relational} and helps the agent correctly judge whether to \textit{proceed} to the next action and \textit{evaluate} the action completions.
%
However, no prior work has systematically studied automatically extracting pre- and postconditions from prevalent data resources. To bridge this gap, we propose the \textit{action condition inference task} on \textbf{real-world instructional manuals}, where a \textbf{\textit{dense} dependency graph} is produced, as in~\figref{fig:teaser}, to denote the pre- and postconditions of actions.
Such a dependency graph provides a systematic task execution plan that agents can closely follow.


We consider two online instruction resources, \textit{WikiHow}~\cite{wikihow} and \textit{Instructables.com}~\cite{instructables}, to study the current NLP models' capabilities of performing the proposed task. 
As there is no densely annotated dataset on the desired action-condition-dependencies from real-world instructions, and annotating a comprehensive dependency structure of actions for long instruction contexts can be extremely expensive and laborious, we collect human annotations on a subset of totally \numdata~samples and benchmark models in either a \textbf{zero-shot} setting where no annotated data is used for training, or a \textbf{low-resource/shot} setting with limited amount of annotated training data.

We also design the following heuristics and show that they can effectively construct 
large-scale \textit{weak supervisions}:
(1) \textbf{Key entity tracing:} Key repetitive entity mentions (including \textbf{co-references}) across different instruction descriptions likely suggest a dependency.
(2) \textbf{Keywords:} Certain keywords (\eg the \underline{before} in ``\textit{do \texttt{X} \underline{before} doing \texttt{Y}}") can often imply the condition dependencies.
(3) \textbf{Temporal reasoning:} We adopt a temporal relation module~\cite{han2021econet} to alleviate the potential inconsistencies between the narrated orders of conditional events and their actual temporal orders to better utilize their temporally grounded nature (\eg preconditions are \textit{prior to} an action).

We benchmark two strong baselines based on pretrained language models with or without instruction contexts on our annotated held-out test-set, where the models are asked to make predictions \textit{exhaustively} on \textbf{every possible dependency}.
We observe that contextualized information is essential (\textgreater~20\% F1-score gain over non-contextualized counterparts), and that our proposed heuristics are able to augment an effective weakly-supervised training data to further improve the performance (\textgreater~6\% F1-score gain) on the low-resource setting.
However, the best results are still well below human performance (\textgreater~20\% F1-score difference).



Our key contributions are three-fold:
(1) We propose an action-condition inference task and create a densely human-annotated \textit{evaluation dataset} to spur research on structural instruction comprehensions.
(2) We design linguistic-centric heuristics utilizing entity tracing, keywords, and temporal reasoning to construct effective large-scale weak supervisions.
(3) We benchmark models on the proposed task to shed lights on future research. 

\section{Terminologies and Problem Definition}
\label{sec:terminologies}

Our goal is to learn to infer action-condition dependencies in real-world instructional manuals.
We first describe essential terminologies in details:

\vspace{.3em}

\mypar{Actionable} refers to a phrase that a person can follow and execute \textit{in the real world} (yellow colored phrases in~\figref{fig:term}).
We also consider negated actions (\eg \textit{do not \underline{...}}) or actions warned to avoid (\eg \textit{if \underline{you purchase the wrong...}}) as they likely also carry useful knowledge regarding the tasks.\footnote{We ask workers to single out the actual \textit{actionable} phrases, \eg \textit{\underline{purchase the wrong line}} $\rightarrow$ \textit{\underline{trimmer will not work}.}}

\vspace{.3em}

\mypar{Precondition} concerns the \textit{prerequisites} to be met for an actionable to be executable, which can be a status, a condition, and/or another prior actionable (blue colored phrases in~\figref{fig:term}).
It is worth noting that humans can omit explicitly writing out certain condition statements because of their triviality as long as the actions inducing them are mentioned (\eg \underline{heat the pan} $\rightarrow$ \underline{pan is heated}, the latter can often be omitted).
We thus generalize the conventional precondition formulation, \ie sets of statements evaluated to true/false~\cite{fikes1971strips}, to a phrase that is either a passive condition statement or an \textit{actionable that induces} the prerequisite conditions, as inspired by~\citet{linden1994generating}.

\vspace{.3em}

\mypar{Postcondition} is defined as the outcome caused by the execution of an actionable, which often involves status changes of certain objects (or the actor itself) or certain effects emerged to the surroundings or world state (green colored phrases in~\figref{fig:term}).

\vspace{.3em}

\mypar{Text segment} in this paper refers to a textual segment of interest, which can be one of: \{actionable, precondition, postcondition\}, in an article.

\vspace{.3em}

\noindent In reality, a valid actionable should have both \textit{pre-} and \textit{postcondition} dependencies,
however, we do not enforce this in this work as conditions can occasionally be omitted by human authors.

\vspace{.3em}

\mypar{Problem Formulation.}
Given an input instructional manual and some text segments of interest extracted from it, a model is asked to predict the \textit{directed} relation between a pair of segments, where the relation should be one of the followings: \texttt{NULL} (no relation), \textit{precondition}, or \textit{postcondition}.

\section{Datasets and Human Annotations}
\label{sec:datasets}

As the condition-dependency knowledge we are interested in is prevalent in real-world instructions, we consider two popular online resources, \textbf{WikiHow} and \textbf{Instructables.com}, both consist of detailed multi-step task instructions, to support our investigation.
For WikiHow, we use the provided dataset from~\citet{wu2022procedural}; for Instructables, we scrape the contents directly from their website.

Since densely annotating large-scale instruction sources for the desired dependencies is extremely expensive and laborious, we mainly annotate a \textit{test-set} and propose to train the models via weakly or self-supervised methods. 
We hence provide a small subset of the human-annotated data to adapt models to the problem domain.
To this end, we collect comprehensive human annotations on a selected subset in each dataset to serve as our \textbf{annotated-set}, and particularly the subsets used to evaluate the models as the \textbf{annotated-test-set}.\footnote{Following~\citet{wu2022procedural}, we first choose from physical categories and then sample a manually inspected subset.}
In total, our densely annotated-set has 500 samples in WikiHow and 150
samples in Instructables, spanning 7,191 distinct actions (defined by main predicate-object phrases) for diversity.
In~\secref{sec:exp_setups}, we will describe how the annotated-set is split to facilitate the low-resource training.
We also collect the human performance on the annotated-test-set to gauge the human upper bound of our proposed task.
More dataset details are in~\appendixref{a-sec:datasets}.

\newcolumntype{M}{>{\centering\arraybackslash}m{.13\textwidth}}
\newcolumntype{N}{>{\arraybackslash}m{.31\textwidth}}

\begin{table*}[t!]{standalone}
\centering
\small
    \begin{tabular}{M|c|N}
    \toprule
    \multicolumn{1}{c|}{\textbf{Heuristics}} & \textbf{Examples} & \multicolumn{1}{c}{\textbf{Descriptions}} \\
    \midrule 
    
    Entity-Tracing \& Coref.
    & \raisebox{-.45\height}{\includegraphics[width=.46\textwidth]{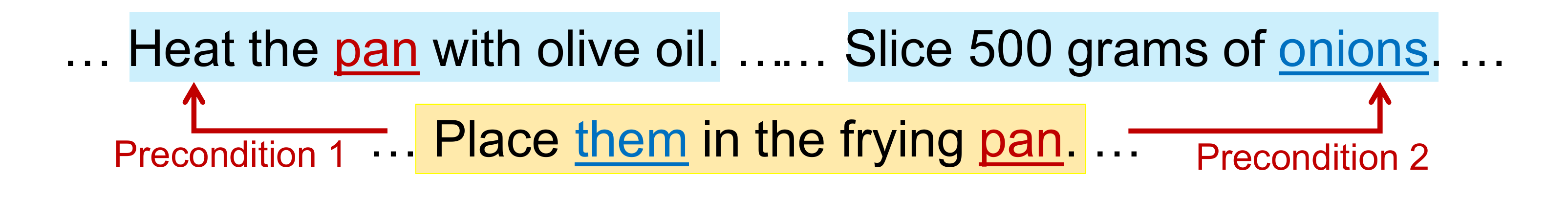}}
    & The shared entities are \underline{pan} and \underline{onions} (linked via co-references to \underline{them}). \\

    \midrule 
    
    Keywords
    & \raisebox{-.5\height}{\includegraphics[width=.48\textwidth]{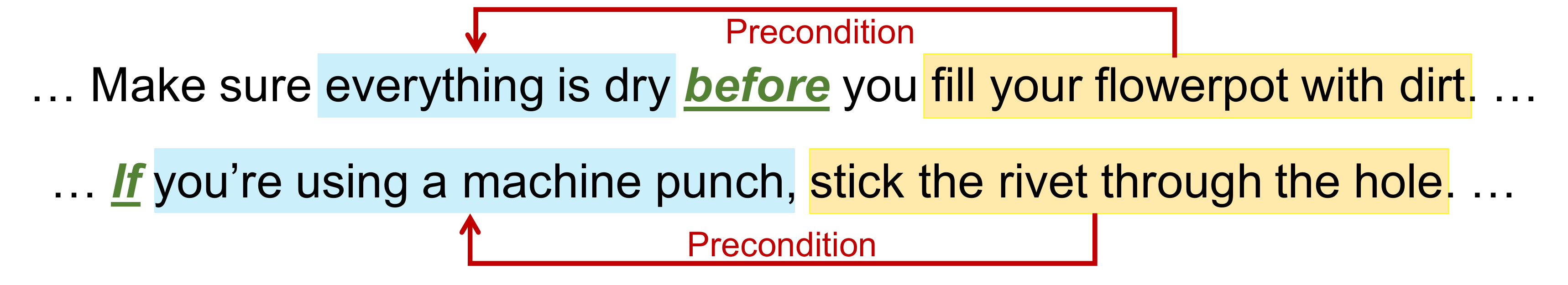}}
    & Keywords are used to link the segments they separate. If the keyword is at the beginning (2nd example), the (1st) comma is used to segment the sentences. \\
    
    \midrule 
    
    Postcondition
    & \raisebox{-.5\height}{\includegraphics[width=.48\textwidth]{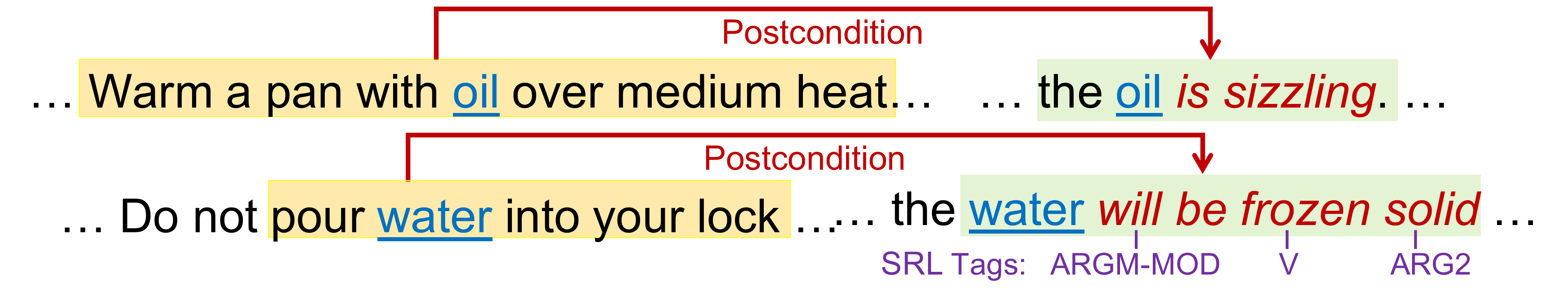}}
    & Certain linguistic hints (\eg SRL tags) are utilized to propose plausible (and likely) postcondition text segments. \\
    
    \midrule 
    
    Temporal
    & \raisebox{-.5\height}{\includegraphics[width=.46\textwidth]{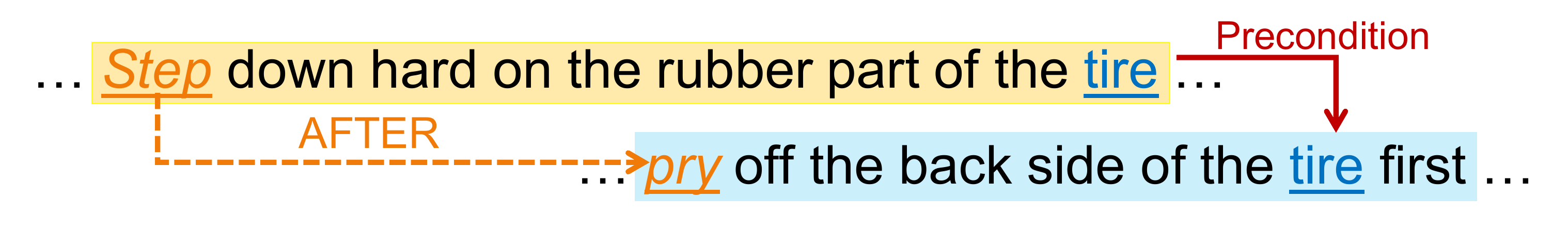}}
    & The action \underline{prying} should occur prior to \underline{stepping}, but these two segments are reversely narrated in the contexts. \\
    
    \bottomrule
    \end{tabular}
\vspace{-.5em}
\caption{
\footnotesize
\textbf{Heuristics} used for determining condition linkages between text segments, with sample use-cases and descriptions.
}
\label{tab:heus_sample_table}
\end{table*}

\subsection{Annotations and Task Specifications}
\label{sec:annot_process}

\mypar{Dataset Structure.}
The desired structure of the constructed data, as in~\figref{fig:term}, features two main components:
(1) \textbf{text segment} of interest (see \secref{sec:terminologies}),
and (2) \textbf{condition linkage}, a \textit{directed} and \textit{relational} link connecting a pair of text segments.

\vspace{.3em}

\mypar{Annotation Process.}
We conduct the annotated-set construction via Amazon Mechanical Turk (MTurk).
Each worker is asked to carefully \textbf{read over thoroughly} a prompted complex multi-step instructional manual,
where the annotation process consists of three main steps:
\textbf{(1) Text segments highlighting:} To facilitate this step (and postulating the text segments for constructing weak-supervisions in~\secref{sec:data_heus}), we \textit{pre-highlight} several text segments extracted by \textit{semantic role labelling} (SRL) for workers to choose from.\footnote{SRL \texttt{V} and \texttt{ARG}s are connected alongside intermediate words to form contiguous segments (see~\appendixref{a-sec:srl_heus}).}
They can also freely annotate (highlight by cursor) their more desirable segments.
\textbf{(2) Linking:} We encourage the workers to annotate all the possible segments of interest, and then they are asked to connect certain pairs of segments that are likely to have dependencies with a directed edge.
\textbf{(3) Labelling:} Finally, each directed edge drawn will need to be labelled as either a \textit{pre-} or \textit{postcondition} (\texttt{NULL} relations do not need to be explicitly annotated).

In general, for each article a worker is required to consider on average $>$$500$ pairwise relations with all associated article contexts ($>$$300$ tokens), which is a \textbf{decently laborious task}.
Comparisons on the linkage annotations from different workers are as well made on \textbf{\textit{every}} pair of \textit{their respective annotated} text segments with the \textbf{\textit{actual} candidate-consideration} from the \textbf{entire} rest of article.

Since the agreements among workers on both text segments and condition linkages are sufficiently high\footnote{The mean inter-annotator agreements (IAAs) per Fleiss Kappa for (segments, linkages) are (0.90, 0.57) and (0.88, 0.58) for WikiHow and Instructables. Note that the Kappa agreement measures the extent to which the observed amount of agreement among raters exceeds what would be expected if all raters made their ratings completely randomly, so the agreement is high. See~\appendixref{a-sec:iaa} for more details.} given the complexity of the annotation task, our final human annotated-set retains the \textit{majority voted} segments and linkages.

\vspace{.3em}

\mypar{Variants of Tasks.}
Although proper machine extraction of the text segments of interest as a span-based prediction can be a valid and interesting task, we find that our automatic SRL extraction is already sufficiently reliable.\footnote{$\sim$58\% of the time SRL-proposed segments were directly used, with others mostly being few-word-span refinements.}
In this paper, we thus mainly focus on the more essential linkage prediction (and their labels) task assuming that these text segments are given, and leave the possible end-to-end system with the (refined) text segment extraction, as the future work.
Our proposed task and the associated annotated-set can be approached by a \textbf{zero-shot} or \textbf{low-resource} setting: the former involves no training on any of the annotated data and a heuristically constructed training set can be utilized (\secref{sec:data_heus}), while the latter allows models to be finetuned on a limited annotated-subset (\secref{sec:multi_staged_training}).
For the low-resource setting particularly, only 30\% of the annotated data will be used for training (details of splits and considerations see~\secref{sec:exp_setups}).


\section{Training With Weak Supervision }
\label{sec:data_heus}


As mentioned in~\secref{sec:datasets}, our proposed task can be approached via a zero-shot setting, where the vast amount of \textbf{un-annotated instruction data} can be transformed into useful training resources (same dataset structure as described in~\secref{sec:annot_process}).
Moreover, it is proven that in many low-resource NLP tasks, constructing a much larger heuristic-based weakly supervised data can be beneficial~\cite{plankagic2018distant,nidhilow}.

\subsection{Linking Heuristics}

The goal of designing certain heuristics is to perform a rule-based determination of the linkage (its direction and the condition label).
Our design intuition is to harness dependency knowledge by exploiting relations between actions and entities (\textit{entity-level}), certain linguistic patterns (\textit{phrase-level}), and \textit{event-level} information, which should be widely applicable to all kinds of instructional data.
Concretely, we design four types of heuristics:
(1) \textbf{Keywords:} certain keywords are hypothesized to show strong implication of conditions such as \textit{if}, \textit{before}, \textit{after};
(2) \textbf{Key entity tracing:} text segments that share the same key entities are likely indicating dependencies;
(3) \textbf{Co-reference} resolution is adopted to supplement (2);
(4) \textbf{Event temporal relation resolution} technique is incorporated to handle the inconsistencies between narrative order and the \textit{actual} temporal order of the events.

\vspace{.2em}

\mypar{SRL Extraction.}
Without access to human refinements (\secref{sec:annot_process}), we leverage SRL to postulate all the segments of interests to construct the weakly-supervised set.
As SRL can detect multiple plausible ways to form the \texttt{ARG} frames with respect to the same \textit{central} verb, we need to additionally determine the most desirable parses \textit{for each action verb}.
In this work, we simply select the most desirable SRL parses by choosing ones that maximize both: (1) the number of plausible segments (each centered around an action verb) \textit{within a sentence}, where they do not overlap above a certain threshold (set to be 60\% in this work), and (2) the number of \texttt{ARG}s in each of such segment.

\subsubsection{Keywords}

\tbref{tab:keywords_details} lists the major keywords that are considered in this work.
Denote a text segment as $a_i$, keywords are utilized so as the text segments separated with respect to them, \ie $a_1$ and $a_2$, can be properly linked.
Different keywords and their positions within sentences can lead to different \textit{directions} of the linkages, \ie $a_1 \rightleftarrows a_2$ (see second row of~\tbref{tab:heus_sample_table}, note that here condition labels are not yet determined).
For example, keywords \underline{before} and \underline{after} intuitively can lead to different directions if they are placed at non-beginning positions.
We follow the rules listed in~\tbref{tab:keywords_details} to decide the directions.

\subsubsection{Key Entity Tracing}
\label{sec:key_entity_tracing}

It is intuitive to assume that if the two text segments mention the same entity, a dependency between them likely exists, and hence a \textit{trace} of the same mentioned entity can postulate potential linkages.
As exemplified in the first row of~\tbref{tab:heus_sample_table}, that \textit{heating the \underline{pan}} being a necessary precondition to \textit{placing onions in the \underline{pan}} can be inferred by the shared mention ``\underline{pan}''.
We adopt two ways to propose the candidate entities:
(1) We extract all the \textit{noun phrases} within the SRL segments (mostly \texttt{ARG}-tags),
(2) Inspired by~\cite{bosselut2017simulating}, a model is learned to predict potential entities involved that are not explicitly mentioned (\eg \textit{\underline{fry the chicken}} may imply a \textit{\underline{pan}} is involved) in the context (more details see~\appendixref{a-sec:entity_trace}).

\vspace{.3em}

\mypar{Co-References.}
Humans often use pronouns to refer to the same entity to alternate the mentions in articles, as exemplified by the mentions \underline{onions} and \underline{them}, in the first row of~\tbref{tab:heus_sample_table}.
Therefore, a straightforward augmentation to the aforementioned entity tracing is incorporating co-references of certain entities.
We utilize a co-reference resolution model~\cite{Lee2018HigherorderCR} to propose possible co-referred terms of extracted entities of each segment within the same step description (we do not consider cross-step co-references for simplicity).

\subsection{Linking Algorithm}
\label{sec:linking_algo}

After applying the aforementioned linking heuristics, each text segment $a_i$, can have $M$ linked segments: \{$a^{l_i}_1, ..., a^{l_i}_M$\}.
For linkages that are \textit{traced} by entity mentions (and co-references), their directions always start from priorly narrated segments to the later ones, while linkages determined by the keywords follow~\tbref{tab:keywords_details} for deciding their directions.
However, the text segments that are narrated too much distant away from $a_i$ are less likely to have direct dependencies.
We therefore \textit{truncate} the linked segments by ensuring any $a^{l_i}_j$ is narrated \textbf{no more than} ``$S$ step'' ahead of $a_i$, where $S$ is empirically chosen to be $2$ in this work.

Despite pruning the traces with the aforementioned design choice $S$ can largely reduce \textit{condition-irrelevant} segments, such heuristic indeed cannot guarantee the included text segments are always dependent with respect to an actionable.
Our goal here is to exploit the generalization ability of language models to \textit{recognize} segments that are most probable conditions by including as many heuristically proposed linkages as possible, where a better strategy on designing the maximum allowed step-wise distance is left as a future work.

\subsubsection{Incorporating Temporal Relations}

As hinted in~\secref{sec:terminologies}, the conditions with respect to an actionable imply their temporal relations.
The direction of an entity-trace-induced linkage is naively determined by the narrated order of text segments within contexts, however, in some circumstances (\eg fourth row in~\tbref{tab:heus_sample_table}), the narrative order can be inconsistent with the actual temporal order of the events.
To alleviate such inconsistency, we apply an event temporal relation prediction model~\cite{han2021econet} (trained on various temporal relation datasets such as~\textit{MATRES}~\cite{ning2018multi}) to fix the linkage directions.\footnote{These do not include linkages decided by the \underline{\textit{keywords}}.}

We train the model on three different random seeds and make them produce a \textit{consensus} prediction, \ie unless all of the models jointly predict a specific relation (\texttt{BEFORE} or \texttt{AFTER}), otherwise the relation will be regarded as \texttt{VAGUE}.
The model is then applied to predict temporal relations
of each pair of event triggers (extracted by SRL, \ie verbs/predicates),
and then we invert the direction of an entity-trace-induced linkage, $a^{l_i}_j$ $\rightarrow$ $a_i$, if their predicted temporal relation is opposite to their narrated order (\texttt{VAGUE} is of course ignored).

\newcolumntype{Q}{>{\centering\arraybackslash}m{.45\columnwidth}}

\begin{table}[t!]
\centering
\small
    \begin{tabular}{Q|cc}
    \toprule
    \multicolumn{1}{c|}{\textbf{Keywords}} & Begin. & Within Sent. \\
    \midrule
    
    \underline{before},
    \underline{until},
    \underline{in order to},
    \underline{so}
    & $a_1 \longrightarrow a_2$ &  $a_1 \longleftarrow a_2$  \\
    
    \midrule
    
    \underline{requires}
    & --- &  $a_1 \longrightarrow a_2$  \\
    
    \midrule
    
    \underline{after},
    \underline{once},
    \underline{if}
    & $a_1 \longleftarrow a_2$ &  $a_1 \longrightarrow a_2$  \\
    \bottomrule
    \end{tabular}
\vspace{-.5em}
\caption{
\footnotesize
\textbf{Keywords for deciding a potential linkage:} If a keyword is at the beginning of a sentence, we use the (first) comma of that sentence to separate it to two segments and link them accordingly, while the keyword itself is used as the separator otherwise. The segments are then either refined with SRL or kept as they are if SRL does not detect a valid verb.
}
\label{tab:keywords_details}
\end{table}

\begin{figure*}[t!]
\centering

\begin{subfigure}[b]{.35\textwidth}
  \centering
  \includegraphics[width=.98\textwidth]{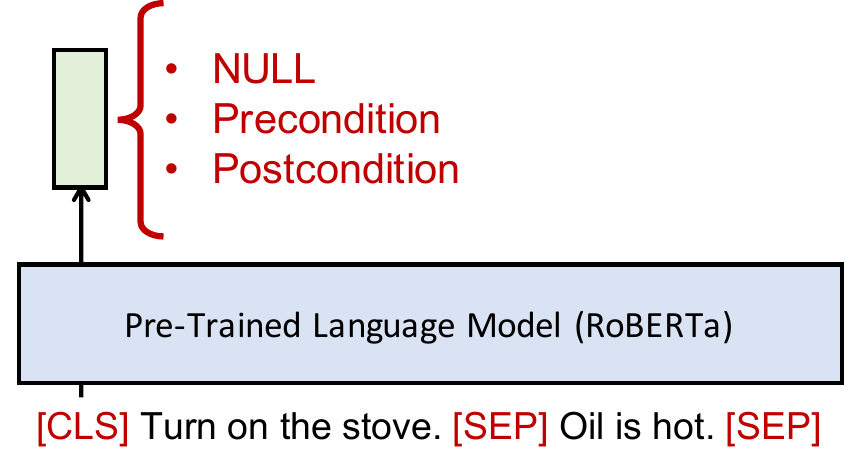}
  \caption{Non-Contextualized Model}
  \label{fig:non_context_model}
\end{subfigure}%
\begin{subfigure}[b]{.63\textwidth}
  \centering
  \includegraphics[width=.98\textwidth]{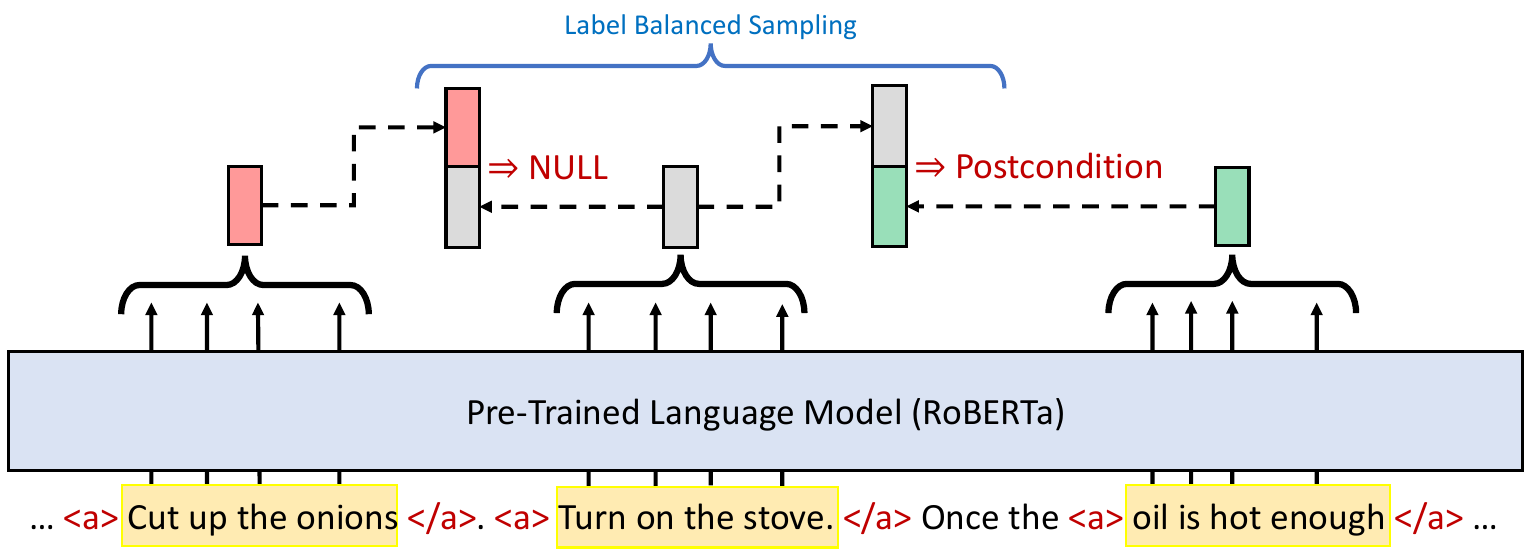}
  \caption{Contextualized Model}
  \label{fig:context_model}
\end{subfigure}
\vspace{-.6em}
\caption{ \footnotesize
\textbf{Model architectures:} 
\textbf{(a) Non-contextualized model:} The model only considers a pair of given text segments.
\textbf{(b) Contextualized model:} The model takes the whole instruction paragraphs (\ie contexts) and wrap each text segment with our special tokens (\texttt{<a>}), where each segment representation is obtained by taking an average over its token representations. The \textit{ordered} concatenated segment representations will then be fed into an MLP to make the final predictions. 
}
\label{fig:model_archis}
\vspace{-.3em}
\end{figure*}

\subsubsection{Labelling The Linkages}

It is rather straightforward to label precondition linkages as a simple heuristic can be used: for a given segment, \textit{any segments that linked to the current one that are either narrated or temporally prior to it} are plausible candidates for being preconditions.
For determining postconditions, where they are mostly descriptions of status (changes), we therefore make use of certain linguistic cues that likely indicate human written status, \eg \textit{the water \underline{will be frozen}} and \textit{the oil \underline{is sizzling}}.
Specifically, we consider:
(1) \textit{be-verbs} followed by \underline{present-progressive tenses} if the subject \textbf{is an entity},
and (2) segments whose SRL tags start with \texttt{ARGM} as exemplified in~\tbref{tab:heus_sample_table}.

\section{Models}

Our proposed heuristics do not assume specific model architecture to be applicable, and to benchmark the proposed task, we mainly consider two types of \textbf{base models}:
(1) \textbf{Non-contextualized} model takes only the \textit{two text segments} of interest at a time and make the \textit{pairwise} trinary (directed) relation predictions, \ie \texttt{NULL}, \textit{precondition}, and \textit{postcondition};
(2) \textbf{Contextualized} model also makes the relation predictions for every pair of input segments, but the inputs include the whole instruction article so the contexts are preserved.
The two models are both based off pretrained language models (the non-contextualized model is essentially a standard transformer-based language model finetuned for classification tasks), and the relation prediction modules are multi-layer perceptrons (MLPs) added on top of the language models' outputs. Cross-entropy loss is used for training.

\subsection{Non-Contextualized Model}

The non-contextualized model takes two separately extracted text segments, $a_{i}$ and $a_{j}$, as inputs and is trained similarly to the next sentence prediction in BERT~\cite{devlin2019bert} (\ie the order of the segments matters, which will be considered in determining their relations), as shown in~\figref{fig:non_context_model}.

\subsection{Contextualized Model}

The architecture of the contextualized model is as depicted in~\figref{fig:context_model}.
Denote the tokens of the instruction text as $\{t_i\}$ and the tokens of $i$-th text segment of interest (either automatically extracted by SRL or annotated by humans) as $\{a_{ij}\}$.
A special start and end of segment token, \texttt{<a>} and \texttt{</a>}, is wrapped around each text segment and hence the input tokens become: "$t_1, ..., t_k, \texttt{<a>}\ a_{i1}, a_{i2}, ..., a_{iK}\ \texttt{</a>}, ...$".
The contextualized segment representation is then obtained by applying a mean pooling over the language model output representations of each of its tokens, \ie denote the output representation of $a_{ij}$ as $\textbf{o}(a_{ij})$, the segment representation of $\textbf{o}(a_{i})$ is $AvgPool(\sum_{j=1}^{K}\textbf{o}(a_{ij}))$.
To determine the relation between segment $i$ and $j$, we feed their \textit{ordered} concatenated representation, $concat(\textbf{o}(a_{i}), \textbf{o}(a_{j}))$, to an MLP for the relation prediction.

\subsection{Learning}
\label{sec:multi_staged_training}

\mypar{Multi-Staged Training.}
For different variants of our task (\secref{sec:annot_process}), we can utilize different combinations of the heuristically constructed dataset and the annotated-train-set.
For the low-resource setting, our models can thus be firstly trained on the constructed training set, and then finetuned on the annotated-set.
Furthermore, following the \textbf{self-training} paradigm~\cite{xie2020self, du2020self}, the previously obtained model predictions can be utilized to either \textit{augment} (\ie adding linkages) or \textit{correct} (\ie revising linkages) the original heuristically constructed data.
And hence a second-stage finetuning can be conducted on this model-self-annotated data for improved performance.

\vspace{.3em}

\mypar{Label Balancing.}
It is obvious that most of the relations between randomly sampled text segment pairs will be \texttt{NULL}, and therefore the training labels are imbalanced.
To alleviate this, we downsample the negative samples when training the models.
Specifically, we fill each training mini-batch with equal amount of positive (relations are not \texttt{NULL}) and negative pairs, where the negatives are constructed by either \textit{inverting} the positive pairs or \textit{replacing} one of the segment with another randomly sampled \textit{unrelated} segment within the same article.

\section{Experiments and Analysis}

Our experiments seek to answer these questions:
(1) How well can the models and humans perform on the proposed task?
(2) Is instructional context information useful?
(3) Are the proposed heuristics and the second-stage self-training effective?

\definecolor{LightCyan}{rgb}{0.88,1,1}
\newcolumntype{a}{>{\columncolor{LightCyan}}d}

\definecolor{LightGreen}{rgb}{0.9,1,0.9}
\newcolumntype{a}{>{\columncolor{LightGreen}}c}

\begin{table*}[th!]
\centering
\small\addtolength{\tabcolsep}{-1.4pt}
\scalebox{.87}{
    \begin{tabular}{ccc|ccc|ccc|ccc|ccc}
    \toprule
    \multicolumn{1}{c}{\multirow{3}{*}{\textbf{Model}}} & \multicolumn{1}{c}{\multirow{3}{*}{\textbf{Heus.}}} & \multicolumn{1}{c|}{\multirow{3}{*}{\textbf{Finetuned/Self}}} & 
    \multicolumn{6}{c|}{\multirow{1}{*}{\textbf{\textcolor{blue}{WikiHow Annotated-Test-Set}}}} &
    \multicolumn{6}{c}{\multirow{1}{*}{\textbf{\textcolor{blue}{Zero-Shot Transfer to Instructables}}}}
    \\
    & & &
    \multicolumn{3}{c|}{\textbf{Precondition}} & \multicolumn{3}{c|}{\textbf{Postcondition}} &
    \multicolumn{3}{c|}{\textbf{Precondition}} & \multicolumn{3}{c}{\textbf{Postcondition}}
    \\
    & & &
    Prec. & Recall & F-1 & Prec. & Recall & \multicolumn{1}{c|}{F-1} &
    Prec. & Recall & F-1 & Prec. & Recall & \multicolumn{1}{c}{F-1} \\
    \midrule
    
    
    \multirow{1}{*}{Prob. Random}
    & \multirow{1}{*}{---} & N/N
      & 3.55 & 4.42 & 3.54 & 0.61 & 0.86 & 0.68
      & 2.94 & 3.88 & 3.04 & 0.46 & 0.46 & 0.42
      \\
    
    \multirow{1}{*}{Prompt. GPT-3}
    & \multirow{1}{*}{---} & N/N
      & 3.87 & 73.46 & 7.35 & 4.90 & \textbf{77.08} & 9.21
      & 3.14 & 64.25 & 5.99 & 1.37 & 34.33 & 2.65
      \\
    
    \multirow{1}{*}{Adapt.-XPAD}
    & \multirow{1}{*}{---} & Y/N
      & 6.21 & 58.38 & 10.64 & 9.47 & 13.83 & 10.45
      & 5.11 & 57.53 & 8.92 & 7.74 & 9.00 & 7.89
      \\
      \midrule
    
    \multirow{2}{*}{Non-Context.}
    & \multirow{1}{*}{Y} & Y/N
      & 8.21 & 79.52 & 14.32 & 15.43 & 44.99 & 20.56
      & 6.49 & 65.05 & 11.31 & 13.64 & 43.50 & 18.65
      \\
    & \multirow{1}{*}{Y} & Y/Y
      & 8.56 & \textbf{81.19} & 14.91 & 26.53 & 65.95 & 34.31
      & 6.64 & \textbf{67.13} & 11.54 & 24.53 & \textbf{61.93} & 31.78
      \\
    \cline{2-15} \\[-.8em]

    \multirow{5}{*}{Context.}
    & \multirow{1}{*}{N} & Y/N
      & 34.01 & 58.33 & 39.27 & 34.44 & 43.15 & 36.79
      & 26.93 & 53.43 & 32.92 & 32.16 & 41.39 & 34.42
      \\
    & \multirow{1}{*}{N} & Y/Y
      & 42.26 & 58.45 & 45.41 & 40.99 & 46.51 & 42.32
      & 38.16 & 55.77 & 42.23 & 42.57 & 48.00 & 44.07
      \\
    \cline{2-15} \\[-.8em]
    & \multirow{1}{*}{Y} & N/N
      & 10.69 & 34.79 & 15.05 & 10.34 & 11.88 & 10.49
      & 10.34 & 16.17 & 11.42 & 4.52 & 4.15 & 4.15
      \\
    & \multirow{1}{*}{Y} & Y/N
      & 47.92 & 64.63 & 51.38 & 51.15 & 57.64 & 52.59
      & 40.70 & 58.97 & 45.17 & 47.92 & 56.51 & 50.06
      \\
    & \multirow{1}{*}{Y} & Y/Y
      & \textbf{49.42} & 68.40 & \textbf{53.51} & \textbf{52.39} & 57.35 & \textbf{53.42}
      & \textbf{43.81} & 62.71 & \textbf{48.34} & \textbf{53.41} & 60.51 & \textbf{55.17}
      \\
    
    \midrule
    \multicolumn{1}{a}{\multirow{1}{*}{Human}}
    & \multicolumn{1}{a}{---} & \multicolumn{1}{a|}{---}
      & \multicolumn{1}{a}{83.91} & \multicolumn{1}{a}{83.86} & \multicolumn{1}{a|}{83.55}
      & \multicolumn{1}{a}{77.39} & \multicolumn{1}{a}{84.81} & \multicolumn{1}{a|}{78.81}
      & \multicolumn{1}{a}{84.74} & \multicolumn{1}{a}{81.32} & \multicolumn{1}{a|}{82.78}
      & \multicolumn{1}{a}{71.90} & \multicolumn{1}{a}{82.51} & \multicolumn{1}{a}{75.53}
      \\

    \bottomrule
    
    \end{tabular}
}
\vspace{-.5em}
\caption{
\footnotesize
\textbf{Annotated-test-set performance:} The best performance is achieved by applying all of the proposed \textbf{heuristics (heus.)} and undergoing the two-stage training: \textbf{finetuned} on the annotated-train-set first and then perform the \textbf{self}-training. Note that for the Instructables, both \textit{Finetuned} and \textit{Self} are done on the WikiHow training sets and a \textbf{zero-shot} transfer is performed.
}
\label{tab:main_results}
\vspace{-.3em}
\end{table*}

\definecolor{LightCyan}{rgb}{0.88,1,1}
\newcolumntype{a}{>{\columncolor{LightCyan}}d}

\definecolor{LightGreen}{rgb}{0.9,1,0.9}
\newcolumntype{a}{>{\columncolor{LightGreen}}c}

\begin{table*}[th!]
\centering
\small\addtolength{\tabcolsep}{-1.3pt}
\scalebox{.93}{
    \begin{tabular}{c|ccc|ccc|ccc|ccc}
    \toprule
    \multicolumn{1}{c|}{\multirow{3}{*}{\textbf{Heuristics.}}} &
    \multicolumn{6}{c|}{\multirow{1}{*}{\textbf{\textcolor{blue}{WikiHow Annotated-Test-Set}}}} &
    \multicolumn{6}{c}{\multirow{1}{*}{\textbf{\textcolor{blue}{Zero-Shot Transfer to Instructables}}}}
    \\
    &
    \multicolumn{3}{c|}{\textbf{Precondition}} & \multicolumn{3}{c|}{\textbf{Postcondition}} &
    \multicolumn{3}{c|}{\textbf{Precondition}} & \multicolumn{3}{c}{\textbf{Postcondition}}
    \\
    &
    Prec. & Recall & F-1 & Prec. & Recall & \multicolumn{1}{c|}{F-1} &
    Prec. & Recall & F-1 & Prec. & Recall & \multicolumn{1}{c}{F-1} \\
    \midrule
    

    \multirow{1}{*}{-- temporal -- coref. - keywords}
      & 45.60 & 61.22 & 48.59 & 43.71 & 47.56 & 44.35
      & 39.35 & 57.03 & 43.49 & 38.45 & 42.96 & 39.39
      \\
    \multirow{1}{*}{-- temporal -- coref.}
      & 43.43 & 64.43 & 48.04 & 46.27 & 51.27 & 47.22
      & 37.06 & 59.95 & 42.56 & 38.41 & 44.54 & 39.83
      \\
    \multirow{1}{*}{-- temporal}
      & 45.83 & 62.48 & 49.17 & 47.72 & 52.70 & 48.81
      & 39.39 & 59.53 & 44.23 & 46.81 & 52.15 & 48.23
      \\

    \bottomrule
    
    \end{tabular}
}
\vspace{-.5em}
\caption{
\footnotesize
\textbf{Heuristics ablations:}
The models used here are \textbf{contextualized} models without the second-stage self-training for both datasets, and "--" indicates exclusion (from using all).
In general, each of the designed heuristics give incremental performance gain to both datasets, where the temporal component is particularly effective in postcondition predictions (compare to~\tbref{tab:main_results}).
}
\label{tab:heus_ablations}
\vspace{-.4em}
\end{table*}

\begin{table}[th!]
\centering
\small
\scalebox{.92}{
    \begin{tabular}{c|ccc|ccc}
    \toprule
    \multicolumn{1}{c|}{\multirow{2}{*}{\textbf{Train}}}
    & \multicolumn{3}{c|}{\textbf{Precondition}} & \multicolumn{3}{c}{\textbf{Postcondition}} \\
    & Prec. & Recall & F-1 & Prec. & Recall & F-1 \\
    \midrule
    
    10\% & 41.34 & 61.71 & 46.06 & 45.24 & 55.56 & 47.95 \\
    
    20\% & 45.60 & 67.55 & 50.78 & 49.30 & 58.02 & 51.62 \\
    
    30\% & 57.38 & 64.46 & 57.53 & 50.49 & 54.57 & 51.09 \\
    
    40\% & 49.61 & 73.09 & 55.14 & 50.45 & 57.77 & 52.27 \\
    
    50\% & 54.27 & 70.89 & 57.84 & 51.35 & 55.85 & 52.23 \\
    
    60\% & 53.21 & 69.36 & 56.42 & 53.68 & 58.09 & 54.46 \\
    
    \bottomrule
    \end{tabular}
}
\vspace{-.8em}
\caption{
\footnotesize
\textbf{Varying annotated-train-set size:} on WikiHow (test-set size is fixed at 30\%).
We use the (best) model trained with all the proposed heuristics and the self-training paradigm.
}
\label{tab:varying_train_results_wikihow}
\end{table}

\subsection{Training and Implementation Details}
\label{sec:impl}

For both non-contextualized and contextualized models, we adopt the pretrained RoBERTa (-large) language model~\cite{liu2019roberta} as the base model.
All the linguistic features, \ie SRL~\cite{Shi2019SimpleBM}, co-references, POS-tags, are extracted using models implemented by AllenNLP~\cite{Gardner2017AllenNLP}.
We truncate the input texts at maximum length of 500
while ensuring all the text segments within this length is preserved completely.

All the models in this work (\ie both pretraining and finetuning) are trained on a single Nvidia A100 (40G RAM) GPU.
The hyperparameters are manually tuned against different datasets, and the checkpoints used for testing are selected by the best performing ones on the held-out development sets.

\subsection{Experimental Setups}
\label{sec:exp_setups}

\mypar{Data Splits.}
The primary benchmark of WikiHow annotated-set is partitioned into \textbf{train (30\%)}, \textbf{development (10\%)}, and \textbf{test (60\%)} set, respectively, resulting in 150, 50, and 300 data samples, for low-resource setting. 
We mainly consider the Instructables annotated-set in a \textbf{zero-shot setting} where we hypothesize the models trained on WikiHow can be well-transferred to it.
For training conducted on the heuristically constructed data, including the second-stage self-training, we use respective held-out development sets to select the checkpoints around performance convergence for finetuning.

\vspace{.3em}

\mypar{Evaluation Metrics.}
We ask the models to predict the relations on \textit{every} pair of text segments in a given instruction, and compute the average precision (Prec.), recall, and F-1 scores separately with respect to each (pre/post) condition labels.

\vspace{.3em}

\mypar{Baselines.}
There is no immediate baseline we are aware of for the proposed action condition inference task.
However, we note that~\citet{dalvi2019everything}'s dependency graph prediction on scientific procedures~\cite{mishra2018tracking} shares high-level similarities to specifically our precondition inference task.
Our non-contextualized model (without the second-stage self-training) with \textit{only} the noun-phrase-based entity tracing heuristic resembles the KB-induced \textit{prior dependency likelihood}, $g_{kb}$, in their proposed XPAD framework.\footnote{With all entity-state-related components excluded (irrelevant to our task) and encoder replaced by RoBERTa model.}

Beside this \textbf{\textit{adapted}-XPAD}, we also evaluate our task with
(1) \textbf{probabilistic random-guess baseline} (random guesses proportional to the training-set label ratio), and (2) \textbf{zero-shot GPT-3}~\cite{brown2020language} where we prompt GPT-3 with exemplar data instances as the task definition (\textbf{contextualized}, see~\appendixref{a-sec:gpt-3} for prompts used). 
These baselines help us to set up a benchmark and justify the challenges our task poses.


\subsection{Experimental Results}
\label{sec:main_results}

\tbref{tab:main_results} left half summarizes both the human and model performance on our standard split (30\% train, 60\% test) of WikiHow annotated-set.
Contextualized model obviously outperforms the non-contextualized counterpart greatly, and all learned models perform well-above random baseline.
Significant improvements on both pre- and postcondition inferences can be noticed when heuristically constructed data is utilized, especially when no second-stage self-training is involved.
The best performance is achieved by \textbf{applying all the heuristics} we design, where further improvements are made by augmenting with second-stage pseudo supervisions.
Similar performance trends can be observed in~\tbref{tab:main_results} right half where a zero-shot transfer from models trained on WikiHow data to Instructables is conducted.

Notice that the zero-shot GPT-3 performs quite poorly compared to our \textit{best low-resource training setting}, and generally worse than our zero-shot contextualized model utilizing only the heuristically constructed data.
We hypothetically attribute the poor performance to both the requirement of exhaustive search of the conditions across the whole manual, and its lacking of complex commonsense reasoning; justifying the effectiveness of our proposed training paradigm and the difficulty of our task.
Nevertheless, there are still \textbf{large rooms} for improvement as the best model falls well-behind human performance (\textgreater 20\% F1-score gap).

\vspace{.3em}

\mypar{Heuristics Ablations.}
\tbref{tab:heus_ablations} features ablation studies on the designed heuristics.
One can observe that keywords are mostly effective on inferring the postconditions, and co-references are significantly beneficial in the Instructables data, which can hypothetically be attributed to the writing style of the datasets (\ie authors of Instructables might use co-referred terms more).
Temporal relation resolution is consistently helpful across pre- and postconditions as well as datasets, suggesting only relying on narrated orders could degenerate the performance.



\subsubsection{Error Analysis.}
While our (best) models perform well on linkages that exhibit similar concepts to the designed heuristics and generalize beyond their surface forms, we are interested in investigating under which situations they are more likely to err.
We therefore sub-sample 10\% of the annotated test-set for manual qualitative inspections and summarize our observations in~\tbref{tab:error_sample_table}.
We find that our models can sometimes \textbf{overfit to certain heuristic} concepts as in~\tbref{tab:error_sample_table} first row (within a food preparation context).
Another improvement the models can enjoy is \textbf{better causal understanding}, which is currently not explicitly handled by our heuristics and can be an interesting future work
(\tbref{tab:error_sample_table} second row, in a biking and cleaning contexts).

\newcolumntype{G}{>{\centering\arraybackslash}m{.135\columnwidth}}
\newcolumntype{H}{>{\arraybackslash}m{.52\columnwidth}}
\newcolumntype{J}{>{\arraybackslash}m{.19\columnwidth}}

\begin{table}[t!]
\centering
\small
    \renewcommand{\arraystretch}{0.5} 
    \begin{tabular}{G|H|J}
    \toprule
    \multicolumn{1}{c|}{\textbf{Type}} & \multicolumn{1}{c|}{\textbf{Example}} & \multicolumn{1}{c}{\textbf{Description}} \\
    \midrule 

    Heus. Overfit
    & \scalebox{.7}{\raisebox{.\height}{\includegraphics[width=.75\columnwidth]{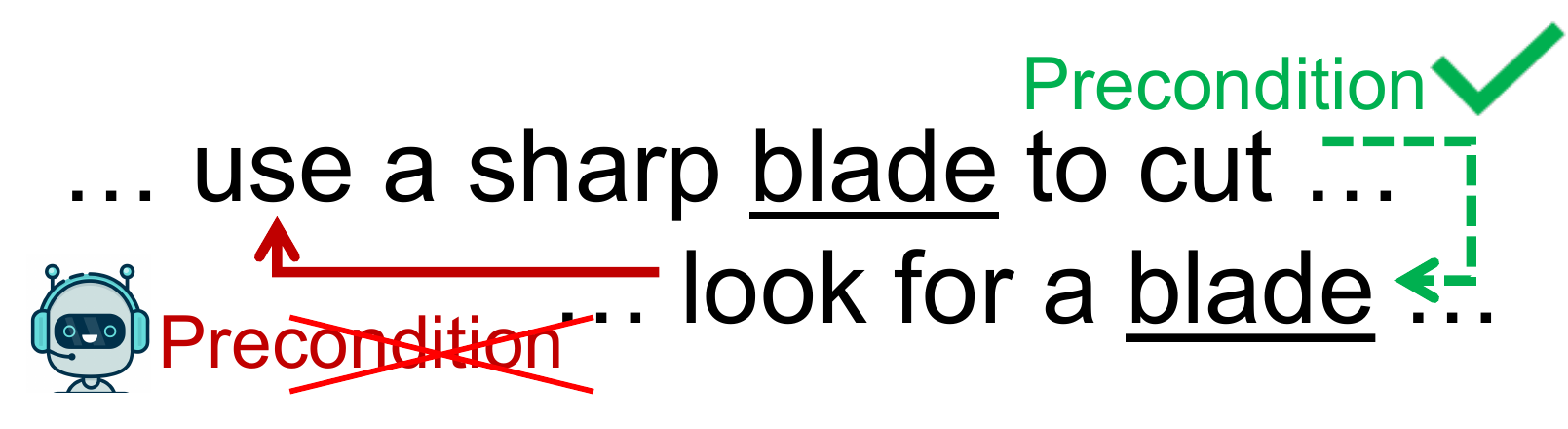}}} & Overfits on entity trace heuristic. \\

    \midrule 
    
    Lacking Causal Reason
    & \scalebox{.7}{\raisebox{.\height}{\includegraphics[width=.78\columnwidth]{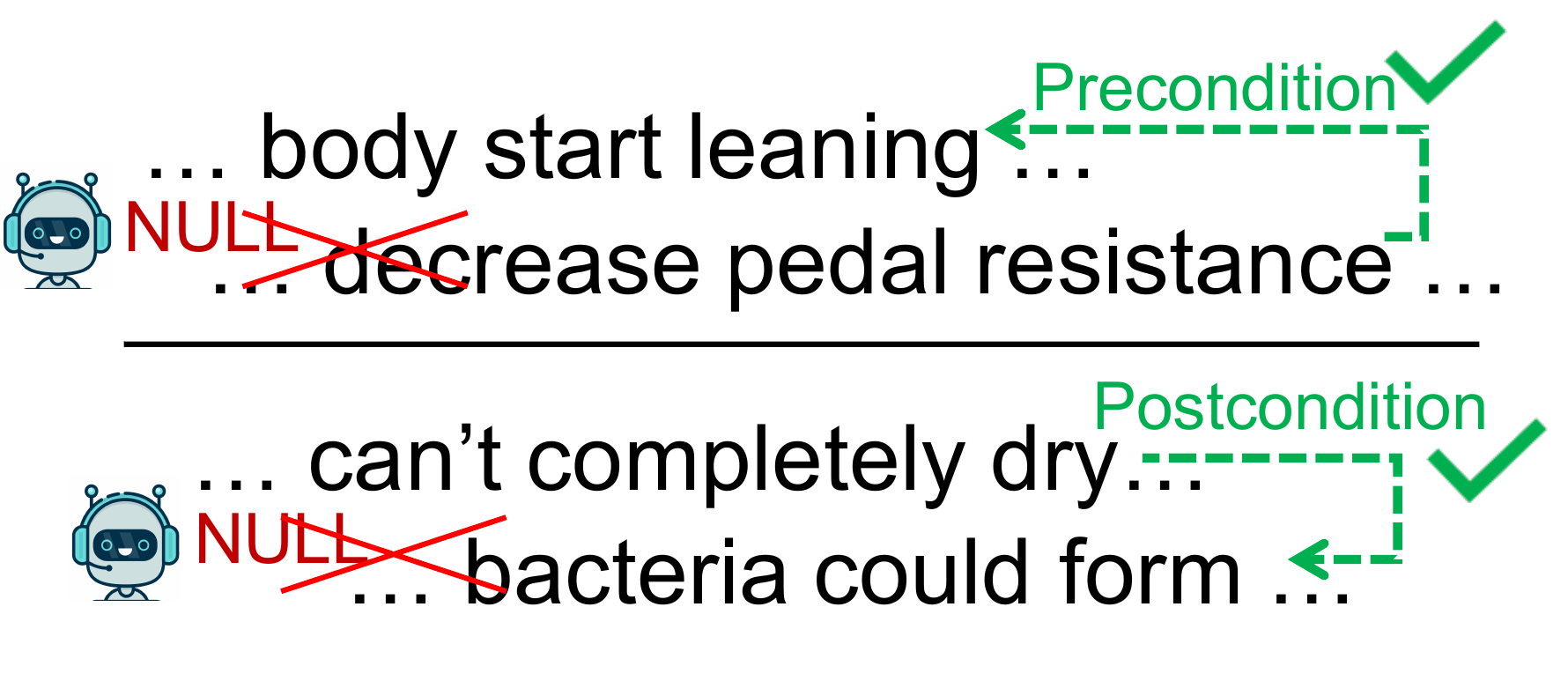}}} & Knowledge-enhanced causal reasoning can be helpful. \\
    
    \bottomrule
    \end{tabular}
\vspace{-.8em}
\caption{
\footnotesize
\textbf{Exemplar model errors.} The second row are from distant segments not link-able even via the keyword heuristic.
}
\label{tab:error_sample_table}
\end{table}

Humans, on the other hand, exhibit much superior performance than the models, tend to fail more often on two kinds of situations:
(1) Missing preconditions (of an action) in those \textit{much earlier paragraphs}, and
(2) Sophisticated temporal ordering of the events (often not narrated sequentially in the texts).
Especially, the first sentences of each task-step are often regarded as the starting actions, while in reality, they can be postconditions of the followed-up detailed contexts.
However, we think both aforementioned errors are rather remediable if the annotators are more careful and search more exhaustively for condition statements.

\subsubsection{The Effect of Training Set Size}
\label{sec:train_set_vary}

\tbref{tab:main_results} shows that with a little amount of data for training, our models can perform significantly better than the zero-shot setting.
This arouses a question -- how would the performance change with respect to the training set size, \ie do we just need more data?
To quantify the effect of training size on model performance, we conduct an experiment where we vary the sample size in the training set while fixing the development (10\%) and test (30\%) set for consistency consideration.
We use the best settings in~\tbref{tab:main_results}, \ie with all the heuristics and self-training paradigm, for this study.
We can observe, from~\tbref{tab:varying_train_results_wikihow}, a plateau in performance when the training set size is approaching 60\%, implying that simply keep adding more training samples does not necessarily yield significant improvements, and hypothesize that the discussed potential improvements are the keys to further effectively exploit the rich knowledge in large-scale instructional data.

\section{Related Works}

\mypar{Procedural Text Understanding.}
Uncovering knowledge in texts that specifically features \textit{procedural structure} has drawn many attentions, including aspects of tracking entity state changes~\cite{branavan-etal-2012-learning,bosselut2017simulating, mishra2018tracking, tandon-etal-2020-dataset}, incorporating common sense or constraints~\cite{tandon2018reasoning,du2019consistent}, procedure-centric question answering (QA)~\cite{tandon2019wiqa}, and structural parsing or generations~\cite{malmaud2014cooking,zellers2021piglet,Zhou2023NonSequentialGS}.
\citet{clark2018happened} leverages VerbNet~\cite{schuler2005verbnet} with \textit{if-then} constructed rules, one of the keywords we also utilize, to determine object-state postconditions for answering state-related reading comprehension questions.
In addition, some prior works also specifically formulate precondition understanding as multiple choice QA
for event triggers (verbs)~\cite{kwon2020modeling} and common sense phrases~\cite{qasemi2021corequisite}.
We hope our work on inferring action-condition dependencies, an essential knowledge especially for understanding task-procedures, from long instruction texts, can help advancing the goal of more comprehensive procedural text understanding.

Drawing dependencies among procedure steps has been explored in~\cite{dalvi2019everything,sakaguchi2021proscript,pal-etal-2021-constructing}, however, their procedures are manually synthesized short paragraphs.
Our work, in contrast, aims at inferring diverse dependency knowledge directly from complex real-world and task-solving-oriented instructional manuals, enabling the condition dependencies to go beyond inter-step and narrative boundaries.


\vspace{.3em}

\mypar{Event Relation Extraction.}
Our work is also inspired by document-level event relation extraction~\cite{han2019joint,han2021ester,huang2021tempgen,ma2021eventplus}.
Specifically, certain works also adopt weak supervisions to learn event temporal relations~\cite{zhou2020temporal,zhou2021temporal,han2021econet}, while other relevant works aim at extracting causality relations (mainly cause-effect) automatically from texts~\cite{cao2016role,altenberg1984causal,stasaski2021automatically}.
Our work combines multiple commonsensical heuristics tailored to the nature of the dependencies exhibited in actions and their conditions, in real-world instruction sources.

\section{Conclusions}

In this work we propose a task on inferring action and (pre/post)condition dependencies on real-world online instructional manuals.
We formulate the problem in both zero-shot and low-resource settings, where several heuristics are designed to construct an effective large-scale weakly supervised data.
While the proposed heuristics and the two-staged training leads to significant performance improvements, the results still highlight significant gaps below human performance (\textgreater~20\% F1-score).

We hope our studies and the collected resources can spur relevant research, and suggest two main future directions:
(1) End-to-end propose (refined) actionables, conditions, and their dependencies, by fully exploiting our featured span-annotations of the text segments.
(2) Inferred world states from the text descriptions as well as external knowledge of the entities and causal common sense can be factored into the heuristics for weak-supervisions.

\clearpage

\section{Limitations}

We hereby discuss the current limitations of our work:
\textbf{(1)} As mentioned in~\secref{sec:annot_process}, although our annotated dataset enables the possibility of learning an extractive model that can be trained to predict the span of the text segments of interest from scratch, we focus on the more essential action-condition dependency linkage inference task as we find that the SRL extraction heuristic currently applied sufficiently reliable.
In the future, we look forward to actualizing such an extractive module and other relevant works that can either further refine the SRL-spans or directly propose the text segments we require.
More specifically, the extractive module can be supervised and/or evaluated against with our human annotations on the text segment start-end positions of an article.
\textbf{(2)} The current system is only trained on unimodal (text-only) and English instruction resources. Multilingual and multimodal versions of our work could be as well an interesting future endeavors to make.
\textbf{(3)} In this work, we mostly consider instructions from physical works. While certain conditions and actions can still be defined within more social domain of data (\eg a precondition to \underline{\textit{being a good person}} might be \underline{\textit{cultivating good habits}}). As a result, we do not really guarantee the performance of our models when applied to data from these less physical-oriented domains.

\section{Ethics and Broader Impacts}

We hereby acknowledge that all of the co-authors of this work are aware of the provided \textit{ACL Code of Ethics} and honor the code of conduct.
This work is mainly about inferring pre- and postconditions of a given action item in an instructional manual.
The followings give the aspects of both our ethical considerations and our potential impacts to the community.

\vspace{.3em}

\mypar{Dataset.}
We collect the human annotation of the ground truth condition-action dependencies via Amazon Mechanical Turk (MTurk) and ensure that all the personal information of the workers involved (e.g., usernames, emails, urls, demographic information, etc.) is discarded in our dataset.
Although we aim at providing a test set that is agreed upon from various people examining the instructions, there might still be unintended biases within the judgements, we make efforts on reducing these biases by collecting diverse set of instructions in order to arrive at a better general consensus on our task.

This research has been reviewed by the \textbf{IRB board} and granted the status of an \textbf{IRB exempt}.
The detailed annotation process (pay per amount of work, guidelines) is included in the appendix; and overall, we ensure our pay per task is above the the annotator's local minimum wage (approximately \$15 USD / Hour).
We primarily consider English speaking regions for our annotations as the task requires certain level of English proficiency.

\vspace{.3em}

\mypar{Techniques.} We benchmark the proposed condition-inferring task with the state-of-the-art large-scale pretrained language models and our proposed training paradigms.
As commonsense and task procedure understanding are of our main focus, we do not anticipate production of harmful outputs, especially towards vulnerable populations, after training (and evaluating) models on our proposed task.
\section*{Acknowledgments}
Many thanks to Rujun Han for his implementation on the temporal relation resolution model.
This material is based on research supported by the Machine Common Sense (MCS) program under Cooperative Agreement N66001-19-2-4032 with the US Defense Advanced Research Projects Agency (DARPA). The views and conclusions contained herein are those of the authors and should not be interpreted as necessarily representing DARPA, or the U.S. Government.

\bibliography{anthology,custom}

\clearpage

\appendix

\section{Details of The Datasets}
\label{a-sec:datasets}

Resource-wise our work utilizes online instructional manuals (\eg WikiHow) following many existing works~\cite{zhou-etal-2019-learning-household,zhang-etal-2020-reasoning,wu2022procedural}, specifically, the large-scale WikiHow training data is provided by~\cite{wu2022procedural}, while we scrape the Instructables.com data on our own. Since Instructables.com dataset tend to have noisier and more free-formed texts, we thus manually sub-sample a smaller (as compared to the test-set of WikiHow) high quality subset.

We report the essential statistics of the annotated-sets in~\tbref{tab:annotated_set_stats}.
Although our definition of actionable is \textbf{any} textual phrase that can be actually \textbf{acted} in the real world, every unique phrase in our dataset is basically a distinct actionable. We compute the number of distinct actions by extracting the main verb-noun phrases (with lemmatization applied) in a text segment as a \textit{valid-action}, and report their counts in ~\tbref{tab:annotated_set_stats} as well. Each unique action in this way can lead to roughly only 1-to-3 pairwise relation instance in our annotated dataset. Both this and the aforementioned unique action count justifies the diversity of our collected annotated-set.

Each unique URL of WikiHow can have different multi-step~\textit{sections}, and we denote each unique section as a \textit{unique article} in our dataset; while for Instructables.com, each URL only maps to a single section.
As a result, for WikiHow we firstly manually select a set of URLs that are judged featuring high quality (\ie articles consisting clear instructed actions, and contain not so much non-meaningful or unhelpful monologues from the writer) instructions and then sample one or two sections from each of the URLs to construct our annotated-set.
The statistics of the datasets used to construct the large-scale weakly supervised WikiHow training set can be found in Section 3 of~\cite{wu2022procedural}, where we use their provided WikiHow training samples that are mostly from physical categories.

$^{*}$Our densely annotated datasets and relevant tools will be made public upon paper acceptance.

\subsection{Dataset Splits}
\label{a-sec:data_splits}

The whole annotated Instructables.com data samples are used as an evaluating set so we do not need to explicitly split them.
For WikiHow, we split mainly with respect to the URLs to ensure that no articles (\ie sections) from the same URL are put into different data splits, so as to prevent model exploiting the writing style and knowledge from the same URL of articles on WikiHow.
The splitting on the URL-level is as well a random split.

\section{Details of Human Annotations}
\label{a-sec:human_annots}

\begin{table}[t!]
\begin{subtable}{\columnwidth}
\centering
\small
\scalebox{0.83}{
\begin{tabular}{lrrrr}
    \toprule
    \multicolumn{1}{c}{\textbf{Type}} & \multicolumn{4}{c}{\textbf{Counts}} \\
    \midrule
    \multicolumn{1}{c}{Total Unique Articles} & \multicolumn{4}{c}{500} \\
    \multicolumn{1}{c}{Total Unique URLs} & \multicolumn{4}{c}{326} \\
    \multicolumn{1}{c}{Annot.-Train / Annot.-Test} & \multicolumn{4}{c}{200 / 300} \\
    \multicolumn{1}{c}{Type-Token Ratio} & \multicolumn{4}{c}{9799 / 173920 = 0.06} \\
    \multicolumn{1}{c}{Pre-/Postcondition Ratio} & \multicolumn{4}{c}{16457 / 2839 = 5.80} \\
    \multicolumn{1}{c}{Distinct Actions} & \multicolumn{4}{c}{5205} \\
    \multicolumn{1}{c}{Avg. Instance per Unique Action} & \multicolumn{4}{c}{3.33} \\
    \multicolumn{1}{c}{Avg. Possible Text Segment Pairs} & \multicolumn{4}{c}{717.49} \\
    \toprule
    \multicolumn{1}{c}{\textbf{Type}} & \multicolumn{1}{c}{\textbf{Mean}} & \multicolumn{1}{c}{\textbf{Std}} & \multicolumn{1}{c}{\textbf{Min}} & \multicolumn{1}{c}{\textbf{Max}} \\
    \midrule
    \multicolumn{1}{l}{Tokens in a Step Text}     & 67.67 & 23.77 & 2 & 161 \\
    \multicolumn{1}{l}{Sentences in a Step Text}  & 4.20 & 1.00 & 1 & 6 \\
    \multicolumn{1}{l}{Tokens in an article}  & 319.12 & 91.71 & 96 & 631 \\    
    \multicolumn{1}{l}{Sentences in an article}  & 19.81 & 4.03 & 11 & 28 \\    
    \bottomrule
\end{tabular}
}
\caption{\footnotesize WikiHow}
\label{tab:annotated_set_stats_wikihow}
\end{subtable}

\vspace{.5em}

\begin{subtable}{\columnwidth}
\centering
\small
\scalebox{0.83}{
\begin{tabular}{lrrrr}
    \toprule
    \multicolumn{1}{c}{\textbf{Type}} & \multicolumn{4}{c}{\textbf{Counts}} \\
    \midrule
    \multicolumn{1}{c}{Total Unique Articles} & \multicolumn{4}{c}{150} \\
    \multicolumn{1}{c}{Total Unique URLs} & \multicolumn{4}{c}{150} \\
    \multicolumn{1}{c}{Annot.-Train / Annot.-Test} & \multicolumn{4}{c}{0 / 150} \\
    \multicolumn{1}{c}{Type-Token Ratio} & \multicolumn{4}{c}{5580 / 60150 = 0.09} \\
    \multicolumn{1}{c}{Pre-/Postcondition Ratio} & \multicolumn{4}{c}{5157 / 698 = 7.39} \\
    \multicolumn{1}{c}{Distinct Actions} & \multicolumn{4}{c}{1986} \\
    \multicolumn{1}{c}{Avg. Instance per Unique Action} & \multicolumn{4}{c}{1.11} \\
    \multicolumn{1}{c}{Avg. Possible Text Segment Pairs} & \multicolumn{4}{c}{633.75} \\
    \toprule
    \multicolumn{1}{c}{\textbf{Type}} & \multicolumn{1}{c}{\textbf{Mean}} & \multicolumn{1}{c}{\textbf{Std}} & \multicolumn{1}{c}{\textbf{Min}} & \multicolumn{1}{c}{\textbf{Max}} \\
    \midrule
    \multicolumn{1}{l}{Tokens in a Step Text}     & 64.75 & 42.57 & 2 & 234 \\
    \multicolumn{1}{l}{Sentences in a Step Text}  & 4.27 & 2.73 & 1 & 17 \\
    \multicolumn{1}{l}{Tokens in an article}  & 333.3 & 143.22 & 124 & 877 \\    
    \multicolumn{1}{l}{Sentences in an article}  & 21.98 & 9.47 & 10 & 50 \\  
    \bottomrule
\end{tabular}
}
\caption{\footnotesize Instructables.com}
\label{tab:annotated_set_stats_instr}
\end{subtable}

\caption{\footnotesize
\textbf{General statistics of the two annotated-sets}: We provide the detailed component counts of the  annotated-sets used in this work, including the statistics of tokens and sentences from the instruction steps (lower halves).
}
\label{tab:annotated_set_stats}
\end{table}

\subsection{Inter-Annotator Agreements (IAAs)}
\label{a-sec:iaa}

There are two types of inter-annotator agreements (IAAs) we compute:
(1) \textbf{IAA on text segments}
and (2) \textbf{IAA on linkages}, and we describe the details of their computations in this section.

\vspace{.3em}

\mypar{IAA on Text Segments.}
For each worker-highlighted text segment, either coming from directly clicking the pre-highlighted segments or their own creations, we compute the percentage of the overlapping of the tokens between segments annotated by different workers. If this percentage is \textgreater~60\% of each segment in comparison, we denote these two segments are \textit{aligned}. Concretely, for all the unique segments of the same article, annotated by different workers, we can postulate a segment dictionary where the \textit{aligned} segments from different worker annotations are combined into the same ones. And hence each worker's annotation can be viewed as a binary existence of each of the items in such a segment dictionary, where we can compute the Cohen's Kappa inter-annotator agreement scores on every pair of annotators to derive the averaged IAA scores.

\vspace{.3em}

\mypar{IAA on Linkages.}
Similar to the construction of a segment dictionary, we also construct a \textit{linkage dictionary} where every link has a \textit{head segment} pointing to the \textit{tail segment}, with both of the segments coming from an item in the segment dictionary. We thus can also treat the annotation of the linkages across different worker annotations as a binary existence and perform similar inter-annotator agreement computations.

The resulting IAAs for each dataset and annotation types are reported in~\secref{sec:annot_process}.

\vspace{.3em}

\mypar{Majority Vote.}
To obtain the final multi-annotator-judged refined data, with our collection budget allowance, we ensure that the number of annotators per data instance (instruction article) is at least 2 (mostly 3), where \textit{consensus} (strict agreement) is used for instances with 2 annotators, and \textit{majority vote} is adopted for 3 annotators.

\subsection{Annotation Process}
\label{a-sec:annots_process}

We adopt Amazon Mechanical Turk (MTurk) to publish and collect our annotations, where each of the annotation in the MTurk is called a Human Intelligence Task (HIT).
As shown in \figref{fig:ui_instr}, on the top of each HIT we have a detailed description of the task's introduction, terminologies, and instructions.
For the terms we define, such as actionables and pre-/postconditions, we also illustrate them with detailed examples.
To make it easier for workers to quickly understand our tasks, we provide a video version explaining important concepts and the basic operations.
We also set up a Frequently Asked Question (FAQ) section and constantly update such section with some questions gathered from the workers.

\figref{fig:ui_ui} shows the layout of the annotation panel.
A few statements are pre-highlighted in grey and each of them is clickable.
These statements are automatically pre-selected using the SRL heuristics described in~\secref{sec:annot_process}, which are supposed to cover as much potential actionables and pre-/postconditions as possible.
Workers can either simply click the pre-highlighted statements or \textit{redo} the selection to get their more desired segments.
The clicked or selected statements will pop up to the right panel as the text-blocks.
For the convenience to manage the page layout, each text-block is \textit{dragable} and can be moved anywhere within the panel.
The workers then should examine with their intelligence and common sense to connect text-blocks (two at a time) by right clicking one of them to \textit{start} a directed linkage (which ends at another text-block) and choose a proper dependency label for that particular drawn linkage.

Since our annotation task can be rather complicated, we would like our workers to fully understand the requirements before proceeding to the actual annotation.
All annotators are expected to pass three qualification rounds, each consisting of 5 HITs, before being selected as an official annotator.
15 HITs are annotated internally in advance as the standard answers to be used to judge the qualification round qualities.

We calculate the IAAs of each annotator against our standard answers to measure their performance in our task.
In each round, only the best performers move on to the next.
At the end of each round, we email annotators to explain the questions they asked or some of the more commonly made mistakes shared across multiple workers.
In total, over 60 workers participated in our task, and 10 of them passed the qualification rounds.

We estimate the time required to complete each of our HITs to be 10-15 minutes, and adjust our pay rate to \$2.5 and \$3 USD for the qualification and the actual production rounds, respectively.
This roughly equates to a \$15 to \$18 USD per hour wage, which is above the local minimum wage for the workers.
We also ensure that each of our data samples in the official rounds is annotated by at least two different \textit{good workers}.

\begin{table}[t!]
\centering
\small
\scalebox{1.0}{
    \begin{tabular}{l|rr}
    \toprule
    \textbf{Confidence Level} & \textbf{WikiHow} & \textbf{Instructables.com} \\
    \midrule 
    5 (Very)       & 27.27  & 16.33 \\
    4 (Fairly)     & 27.11  & 23.47 \\
    3 (Moderately) & 28.25  &  22.95 \\
    2 (Somewhat)   &  16.23  &  29.10 \\
    1 (Not-At-All) &  1.14  &  8.16 \\
    \bottomrule
    \end{tabular}
}
\caption{\footnotesize
\textbf{Confidence-Level Statistics (\%):} In WikiHow, majority (\textgreater~80\%) of the annotators indicate at least~\textgreater~3 (Moderately) confidence level. As for Instructables.com, it has lower confidence level as the articles tend to be more free-formed and noisy, however, there are still more than 60\% of the time workers report confidence levels at least moderately.
}
\label{tab:confidence_level}
\end{table}

\vspace{.3em}

\mypar{Confidence Levels.}
We compute the averaged percentage of confidence levels reported by the workers in~\tbref{tab:confidence_level}.
Note that majority of the workers indicate a \textit{moderately} or \textit{fairly} confidence levels, implying they are sufficiently confident about their annotations.
We also see feedback from workers that some of them rarely use strong words such as \textit{very} to indicate their confidence levels, and hence the resulted statistics of their confidences could be a bit biased towards the medium.

\vspace{.3em}

\mypar{Human Performance.}
We randomly select 100 samples from the WikiHow annotated-test-set and 50 samples from the Instructables.com annotated-test-set for computing the human performance.
The allowed inputs are exactly the same as what models take, \ie given all the instruction paragraph as context and highlighted (postulated text segment boxes) text segments of interests, workers are asked to predict the relations among such segments so as to induce a complete dependency graph.
For each sample, we collect inputs from two different workers, and ensure that the workers are not the ones that give the original annotations of the action-condition dependencies.
The human performance is then computed by taking the averaged metrics similar to the models on the given samples.

\section{Modelling Details}

\subsection{More on Heuristics}
\label{a-sec:more_heus}

\subsubsection{SRL Extraction}
\label{a-sec:srl_heus}

As SRL can detect multiple plausible ways to form the \texttt{ARG} frames to the same \textit{central} verb, we need to determine which one is the most likely to be desirable.
When such multiple argument patterns exist for the same central verb, we simply determine the most desirable formation of segments by maximizing both the number of plausible segments (where they do not overlap above certain threshold, which is set to be 60\% in this work) \textit{within a sentence} and the number of \texttt{ARG}s in each segment.

\subsubsection{Linking Algorithm}
\label{a-sec:linking_algo}

In~\secref{sec:linking_algo} we mention that a maximum distance of $2$ steps between linked segments is imposed to filter out possible non-dependent conditions.
While this still can potentially include many not-so-much depended text segments, our goal is to exploit the generalization ability of large-scale pretrained language models to \textit{recognize} segments that are most probable conditions by including as much as heuristically proposed linkages as possible, which is empirically proven effective.
A better strategy on making such a design choice of maximum allowed step-wise distance is left as a future work.

\subsubsection{Keywords}

About 3\% of the entire un-annotated data have sentences containing the keywords we use in this work (\tbref{tab:keywords_details}). Despite the relatively small amount compared to other heuristics, they are quite effective judging from the results reported in~\tbref{tab:main_results}.

\subsubsection{Key Entity Tracing}
\label{a-sec:entity_trace}

For the key entity tracing heuristic described in~\secref{sec:key_entity_tracing}, as long as two segments share at least one mentioned entity, they can be linked (\ie \textit{traced} by the shared entity).
We do not constraint the number of key entities within a segment, so there can be more than one being used to conduct the tracing.

\vspace{.3em}

\mypar{Constructing Entity Prediction Datasets.}
As mentioned in~\secref{sec:key_entity_tracing}, one way to postulate the key entities is via constructing a predictive model for outputting potentially involved entities.
To do so, we firstly construct an \textit{entity vocabulary} by extracting all the noun phrases within each SRL extracted segments of the entire un-annotated-set articles.
To prevent from obtaining a too much large vocabulary as well as improbable entities, we only retain entities (without lemmatization) that appear with \textgreater~$5$ occurrences in at least one article.

We then train a language model (based on RoBERTa-large as well) where the output is the multi-label multi-class classification results on the predicted entities.
When predicting the key entities for a given segment, we further constraint the predictions to be within the local vocabulary (more than $5$ occurrences) within the article such segment belongs to. 
This model is inspired by the entity selector module proposed in~\cite{bosselut2017simulating} while we only consider single step statements.
We verify the performance of the learned model on the dataset provided by~\cite{bosselut2017simulating} (the entity selection task), where our model can achieve roughly 60\% on F-1 metric, indicating the trained model is sufficiently reliable.

\subsubsection{Temporal Relations}

We use the temporal relation resolution model from~\cite{han2021econet} that is trained on various temporal relation datasets such as~\textit{MATRES}~\cite{ning2018multi}.
We train the model on three different random seeds and make them produce a \textit{consensus} prediction, \ie unless all of the models jointly predict a specific relation (\texttt{BEFORE} or \texttt{AFTER}), otherwise the relation will be regarded as \texttt{VAGUE}.

\subsection{GPT-3 Baseline}
\label{a-sec:gpt-3}

We use the most powerful version of GPT-3 (Davinci)\footnote{https://openai.com/api/pricing/} provided by the OpenAI GPT-3 API (zero-shot prompted version) with the following prompt:

\textit{Extract the preconditions and postconditions from this text:}

\textit{Text: "Slice 500 grams of onion. Heat the pan with olive oil. Wait until the oil is sizzling. Place onions in the frying pan. Stir the onions. In a few minutes, they should be caramelized."}

\textit{Segment 1: "Heat the pan with olive oil."}

\textit{Segment 2: "oil is sizzling."}

\textit{Label: post-condition}

\textit{Text: "Slice 500 grams of onion. Heat the pan with olive oil. Wait until the oil is sizzling. Place onions in the frying pan. Stir the onions. In a few minutes, they should be caramelized."}

\textit{Segment 1: "Slice 500 grams of onion."}

\textit{Segment 2: "Place the onions in the frying pan."}

\textit{Label: pre-condition}

\textit{Text: "Slice 500 grams of onion. Heat the pan with olive oil. Wait until the oil is sizzling. Place onions in the frying pan. Stir the onions. In a few minutes, they should be caramelized."}

\textit{Segment 1: "Slice 500 grams of onion."}

\textit{Segment 2: "Heat the pan with olive oil."}

\textit{Label: no relation}

\textit{Text: "}\texttt{Fill-In an Article}\textit{"}

\textit{Segment 1: "}\texttt{Fill-In Text Segment 1}\textit{"}

\textit{Segment 2: "}\texttt{Fill-In Text Segment 2}\textit{"}

\textit{Label: }\texttt{GPT-3 Prediction}

\noindent In other words, we provide an exemplar simplified instance to instruct what pre- and postconditions should be like to the model with the article context and a pair of text segments of interest.
And then, the GPT-3 model should \textit{generate} the text description-based prediction label (non-case-sensitive). For preconditions we allow verbalized label to be within \{\textit{precondition, pre-condition}\}, and postconditions within \{\textit{postcondition, post-condition}\}. For the \texttt{NULL} relation, we allow  \{\textit{no relation, unrelated, null, none}\}.

\subsection{Development Set Performance}

We select the model checkpoints to be evaluated using the held-out development split (annotated-dev-set).
We also report the performance on this annotated-dev-set in~\tbref{tab:main_dev_results_wikihow}.

\definecolor{LightCyan}{rgb}{0.88,1,1}
\newcolumntype{a}{>{\columncolor{LightCyan}}c}

\begin{table*}[th!]
\centering
\small
\scalebox{.93}{
    \begin{tabular}{crcc|ccc|ccc}
    \toprule
    \multicolumn{4}{c|}{\multirow{1}{*}{\textbf{\textcolor{blue}{WikiHow Annotated-Dev-Set}}}} &
    \multicolumn{3}{c|}{\textbf{Precondition}} & \multicolumn{3}{c}{\textbf{Postcondition}} \\
    \multicolumn{1}{c}{\multirow{1}{*}{\textbf{Model}}} & \multicolumn{1}{c}{\multirow{1}{*}{\textbf{Heuristics}}} & \multicolumn{1}{c}{\multirow{1}{*}{\textbf{Finetuned}}} & \multicolumn{1}{c|}{\multirow{1}{*}{\textbf{Self}}} & 
    Prec. & Recall & F-1 & Prec. & Recall & F-1 \\
    \midrule
    
    
    \multirow{1}{*}{Non-Context.}
    & \multirow{1}{*}{All} & Y & Y
      & 8.22 & 74.77 & 14.00 & 19.70 & 69.94 & 28.36 \\
      
    \midrule

    \multirow{8}{*}{Context.}
    & \multirow{1}{*}{No Heuristics} & Y & N
      & 29.96 & 56.91 & 35.41 & 30.28 & 39.10 & 32.03 \\
    & \multirow{1}{*}{No Heuristics} & Y & Y
      & 40.09 & 57.60 & 43.20 & 41.10 & 48.59 & 42.53 \\
    \cline{2-10} \\[-.8em]
    & \multirow{1}{*}{All} & N & N
      & 9.59 & 32.69 & 13.35 & 7.48 & 9.26 & 7.81 \\
    & \multirow{1}{*}{-- temporal -- coref. - keywords} & Y & N
      & 43.59 & 58.74 & 45.95 & 39.33 & 44.45 & 40.64 \\
    & \multirow{1}{*}{-- temporal -- coref.} & Y & N
      & 38.43 & 60.48 & 42.83 & 39.72 & 47.80 & 41.92 \\
    & \multirow{1}{*}{-- temporal} & Y & N
      & 41.19 & 57.06 & 43.92 & 47.63 & 54.69 & 48.91 \\
    & \multirow{1}{*}{All} & Y & N
      & 45.05 & 59.59 & 47.35 & 45.65 & 50.35 & 46.42 \\
    & \multirow{1}{*}{All} & Y & Y
      & 44.93 & 65.25 & 49.12 & 46.06 & 52.04 & 47.21 \\

    \bottomrule
    
    \end{tabular}
}
\caption{
\footnotesize
\textbf{Annotated-dev-set performance on WikiHow:} Similar to~\tbref{tab:main_results}, we report the development set performance on the WikiHow dataset (Instructables.com does not have the development set as we are conducting a zero-shot transfer).
}
\label{tab:main_dev_results_wikihow}
\end{table*}

\begin{table}[th!]
\centering
\small
\scalebox{.92}{
    \begin{tabular}{c|ccc|ccc}
    \toprule
    \multicolumn{1}{c|}{\multirow{2}{*}{\textbf{Train}}}
    & \multicolumn{3}{c|}{\textbf{Precondition}} & \multicolumn{3}{c}{\textbf{Postcondition}} \\
    & Prec. & Recall & F-1 & Prec. & Recall & F-1 \\
    \midrule
    
    10\% & 33.44 & 56.41 & 38.69 & 42.37 & 53.86 & 45.25 \\
    
    20\% & 35.05 & 60.97 & 40.86 & 40.76 & 51.35 & 43.19 \\
    
    30\% & 44.57 & 60.19 & 47.68 & 43.00 & 47.26 & 43.83 \\
    
    40\% & 39.38 & 72.23 & 46.63 & 45.51 & 54.27 & 47.57 \\
    
    50\% & 40.97 & 69.70 & 47.24 & 49.15 & 59.04 & 51.76 \\
    
    60\% & 46.99 & 71.14 & 52.27 & 48.80 & 56.51 & 50.74 \\
    
    \bottomrule
    \end{tabular}
}
\caption{
\footnotesize
\textbf{Varying annotated-train-set size without weakly supervised training:} on WikiHow (test-set size is fixed at 30\%).
The model used in this experiment is without training on any of the heuristically constructed dataset, but we apply the self-training paradigm.
}
\label{tab:varying_train_results_wikihow_no_heus}
\end{table}

\begin{table}[th!]
\centering
\small
\scalebox{.92}{
    \begin{tabular}{c|ccc|ccc}
    \toprule
    \multicolumn{1}{c|}{\multirow{2}{*}{\textbf{Train}}}
    & \multicolumn{3}{c|}{\textbf{Precondition}} & \multicolumn{3}{c}{\textbf{Postcondition}} \\
    & Prec. & Recall & F-1 & Prec. & Recall & F-1 \\
    \midrule
    
    10\% & 32.25 & 50.50 & 36.36 & 41.37 & 51.37 & 44.03 \\
    
    20\% & 35.95 & 56.99 & 40.89 & 48.77 & 60.10 & 51.86 \\
    
    40\% & 39.62 & 64.19 & 45.77 & 48.83 & 60.30 & 52.08 \\
    
    50\% & 57.38 & 64.46 & 57.53 & 50.49 & 54.57 & 51.09 \\
    
    60\% & 45.62 & 61.02 & 49.06 & 55.00 & 65.04 & 57.54 \\
    
    \midrule
    
    10\% & 27.50 & 50.32 & 32.74 & 34.99 & 47.66 & 38.18 \\
    
    20\% & 26.86 & 51.73 & 32.34 & 40.31 & 52.89 & 43.43 \\
    
    40\% & 30.58 & 64.38 & 38.16 & 44.78 & 60.86 & 49.28 \\
    
    50\% & 39.65 & 63.28 & 45.41 & 50.96 & 59.98 & 53.54 \\
    
    60\% & 39.90 & 65.68 & 45.95 & 49.64 & 58.83 & 51.97 \\
    
    \bottomrule
    \end{tabular}
}
\caption{
\footnotesize
\textbf{Varying annotated-train-set size:} on Instructables.com (test-set size is fixed at 100\%).
Note that here the train-set size is from WikiHow annotated-set, and the 30\% is basically~\tbref{tab:main_results}.
The upper half is with models that utilize both the heuristically constructed dataset and the self-training paradigm, while the lower half is with models that do not use any weak supervisions.
}
\label{tab:varying_train_results_instructables}
\end{table}

\begin{table}[th!]
\centering
\small
\scalebox{.92}{
    \begin{tabular}{c|ccc|ccc}
    \toprule
    \multicolumn{1}{c|}{\multirow{2}{*}{\textbf{Train}}}
    & \multicolumn{3}{c|}{\textbf{Precondition}} & \multicolumn{3}{c}{\textbf{Postcondition}} \\
    & Prec. & Recall & F-1 & Prec. & Recall & F-1 \\
    \midrule
    
    10\% & 39.77 & 61.58 & 44.65 & 45.76 & 53.42 & 47.57 \\
    
    20\% & 42.75 & 64.32 & 47.40 & 47.97 & 56.99 & 50.21 \\
    
    30\% & 52.37 & 64.59 & 54.43 & 50.70 & 55.93 & 51.87 \\
    
    40\% & 43.77 & 68.58 & 49.28 & 45.47 & 53.78 & 47.48 \\
    
    50\% & 51.98 & 67.29 & 54.94 & 50.45 & 54.84 & 51.21 \\
    
    60\% & 47.96 & 69.77 & 52.61 & 47.81 & 52.27 & 48.77 \\

    \midrule
    
    10\% & 26.37 & 51.61 & 31.80 & 31.52 & 47.68 & 35.33 \\
    
    20\% & 28.62 & 56.40 & 34.53 & 33.68 & 48.10 & 37.30 \\
    
    30\% & 37.20 & 60.09 & 42.32 & 37.44 & 45.52 & 39.39 \\
    
    40\% & 32.74 & 68.97 & 40.57 & 36.33 & 47.00 & 39.00 \\
    
    50\% & 40.30 & 65.62 & 45.94 & 44.86 & 53.36 & 46.85 \\
    
    60\% & 38.80 & 68.16 & 45.27 & 42.03 & 51.96 & 44.43 \\
    
    \bottomrule
    \end{tabular}
}
\caption{
\footnotesize
\textbf{Varying annotated-train-set size:} on WikiHow (test-set size is fixed at 30\%).
The upper half is with models that utilize the heuristically constructed dataset, while the lower half is with models that do not use any weak supervisions. Both upper and lower halves do \textbf{not} undergo any second-stage self-training.
}
\label{tab:varying_train_results_wikihow_no_self}
\end{table}

\begin{table}[th!]
\centering
\small
\scalebox{.92}{
    \begin{tabular}{c|ccc|ccc}
    \toprule
    \multicolumn{1}{c|}{\multirow{2}{*}{\textbf{Train}}}
    & \multicolumn{3}{c|}{\textbf{Precondition}} & \multicolumn{3}{c}{\textbf{Postcondition}} \\
    & Prec. & Recall & F-1 & Prec. & Recall & F-1 \\
    \midrule
    
    10\% & 29.59 & 52.25 & 34.76 & 40.31 & 50.26 & 42.92 \\
    
    20\% & 31.46 & 53.34 & 36.37 & 44.11 & 55.32 & 46.94 \\
    
    40\% & 34.02 & 60.66 & 40.20 & 43.62 & 51.56 & 45.43 \\
    
    50\% & 42.57 & 59.24 & 46.38 & 49.83 & 57.26 & 51.77 \\
    
    60\% & 37.69 & 61.36 & 43.34 & 48.49 & 54.29 & 49.70 \\

    \midrule
    
    10\% & 18.44 & 41.85 & 23.20 & 21.97 & 39.08 & 26.02 \\
    
    20\% & 20.91 & 48.63 & 26.52 & 28.93 & 44.85 & 32.98 \\
    
    40\% & 23.89 & 61.51 & 31.59 & 36.43 & 51.98 & 40.50 \\
    
    50\% & 30.56 & 58.10 & 36.90 & 41.35 & 54.48 & 44.95 \\
    
    60\% & 28.59 & 60.24 & 35.52 & 40.06 & 53.41 & 43.20 \\
    
    \bottomrule
    \end{tabular}
}
\caption{
\footnotesize
\textbf{Varying annotated-train-set size:} on Instructables.com (test-set size is fixed at 100\%).
The structure of this table is similar to that of~\tbref{tab:varying_train_results_wikihow_no_self}, \ie no self-training is conducted.
}
\label{tab:varying_train_results_instructables_no_self}
\end{table}

\subsection{More Results on Train-Set Size Varying}
\label{a-sec:train_set_vary}

\tbref{tab:varying_train_results_wikihow_no_heus} is a similar experiment as~\tbref{tab:varying_train_results_wikihow} but here we conduct the experiments with the models that do not utilize the weakly supervised data constructed with the proposed heuristics at all.
One can observe that similar trends hold that a plateau can be noticed when the training set size is approaching 60\%.
Compared to~\tbref{tab:varying_train_results_wikihow}, we can also observe that the smaller the train-set size is, the larger gaps shown between the models with and without utilizing the heuristically constructed data. This can further imply the effectiveness of our heuristics to construct meaningful data for the action-condition dependency inferring task.
The models with heuristics, if compared at the same train-set size respectively, significantly outperforms every model counterparts that do not utilize the heuristics.

\tbref{tab:varying_train_results_instructables} reports similar experiments but in the Instructables.com annotated-test-set. Note that we perform a direct zero-shot transfer from the WikiHow annotated-train-set, so the test-set size is always 100\% for the Instructables.

Finally, both \Cref{tab:varying_train_results_wikihow_no_self,tab:varying_train_results_instructables_no_self} report the same experiments, however, this time the second-stage self-training is not applied.
It is worth noting that the self-training is indeed effective throughout all the train-set-size and across different datasets and model variants, however, the trends of model performance hitting a saturation point when the train-set size increases still hold.

\subsection{Training \& Implementation Details}
\label{a-sec:impl}

\begin{table*}[t]
\centering
\footnotesize
\scalebox{0.9}{
    \begin{tabular}{lccccccc}
    \toprule
    & \multicolumn{1}{c}{\multirow{2}{*}{\textbf{Models}}} & \multirow{2}{*}{\textbf{Batch Size}} & \multirow{2}{*}{\textbf{Initial LR}} & \multirow{2}{*}{\textbf{\# Training Epochs}} & \textbf{Gradient Accu-} & \multirow{2}{*}{\textbf{\# Params}}  \\
    & & & & & \textbf{mulation Steps} & \\
    \midrule
    & Non-contextualized  & 8 & $1 \times 10^{-5}$ & 15  & 1 & 355M \\
    & Contextualized      & 4 & $1 \times 10^{-5}$ & 15  & 1 & 372M \\
    \bottomrule
    \end{tabular}
}
\caption{\footnotesize
\textbf{Hyperparameters in this work:} \textit{Initial LR} denotes the initial learning rate. All the models are trained with Adam optimizers~\cite{kingma2014adam}. We include number of learnable parameters of each model in the column of \textit{\# params}.
}
\label{tab:hyparams}
\end{table*}

\begin{table*}[t]
\centering
\footnotesize
\begin{tabular}{ccccc}
    \toprule
    \textbf{Type} & \textbf{Batch Size} & \textbf{Initial LR} & \textbf{\# Training Epochs} & \textbf{Gradient Accumulation Steps} \\
    \midrule
    \textbf{Bound (lower--upper)} & 2--8 & $1 \times 10^{-5}$--$1 \times 10^{-6}$ & 5--15 & 1 \\
    \midrule
    \textbf{Number of Trials} & 2--4 & 2--3 & 2--4 & 1 \\
    \bottomrule
\end{tabular}
\caption{\footnotesize
\textbf{Search bounds} for the hyperparameters of all the models.
}
\label{tab:search}
\end{table*}

\mypar{Training Details.}
The maximum of 500 token length described in~\secref{sec:impl} is sufficient for most of the data in the annotated-test-sets, as evident in~\tbref{tab:annotated_set_stats}.
All the models in this work are trained on a single Nvidia A100 GPU\footnote{https://www.nvidia.com/en-us/data-center/a100/} on a Ubuntu 20.04.2 operating system.
The hyperparameters for each model are manually tuned against different datasets, and the checkpoints used for testing are selected by the best performing ones on the held-out development sets in their respective datasets.

\vspace{.3em}

\mypar{Implementation Details.}
The implementations of the transformer-based models are extended from the HuggingFace\footnote{https://github.com/huggingface/transformers}~code base~\cite{wolf-etal-2020-transformers}, and our entire code-base is implemented in PyTorch.\footnote{https://pytorch.org/}

\subsection{Hyperparameters}

We train our models until performance convergence is observed on the heuristically constructed dataset.
The training time for the weakly supervised learning is roughly 6-8 hours.
For all the finetuning that involves our annotated-sets, we train the models for roughly 10-15 epochs for all the model variants, where the training time varies from 1-2 hours.
We list all the hyperparameters used in~\tbref{tab:hyparams}.
The basic hyperparameters such as learning rate, batch size, and gradient accumulation steps are kept consistent for all kinds of training in this work, including training on the weakly supervised data, finetuning on the annotated-sets, as well as during the second-stage self-training.
All of our models adopt the same search bounds and ranges of trials as in~\tbref{tab:search}.

\begin{figure*}[t!]

\begin{subtable}{\textwidth}
\centering
\centering
    \includegraphics[width=.95\columnwidth]{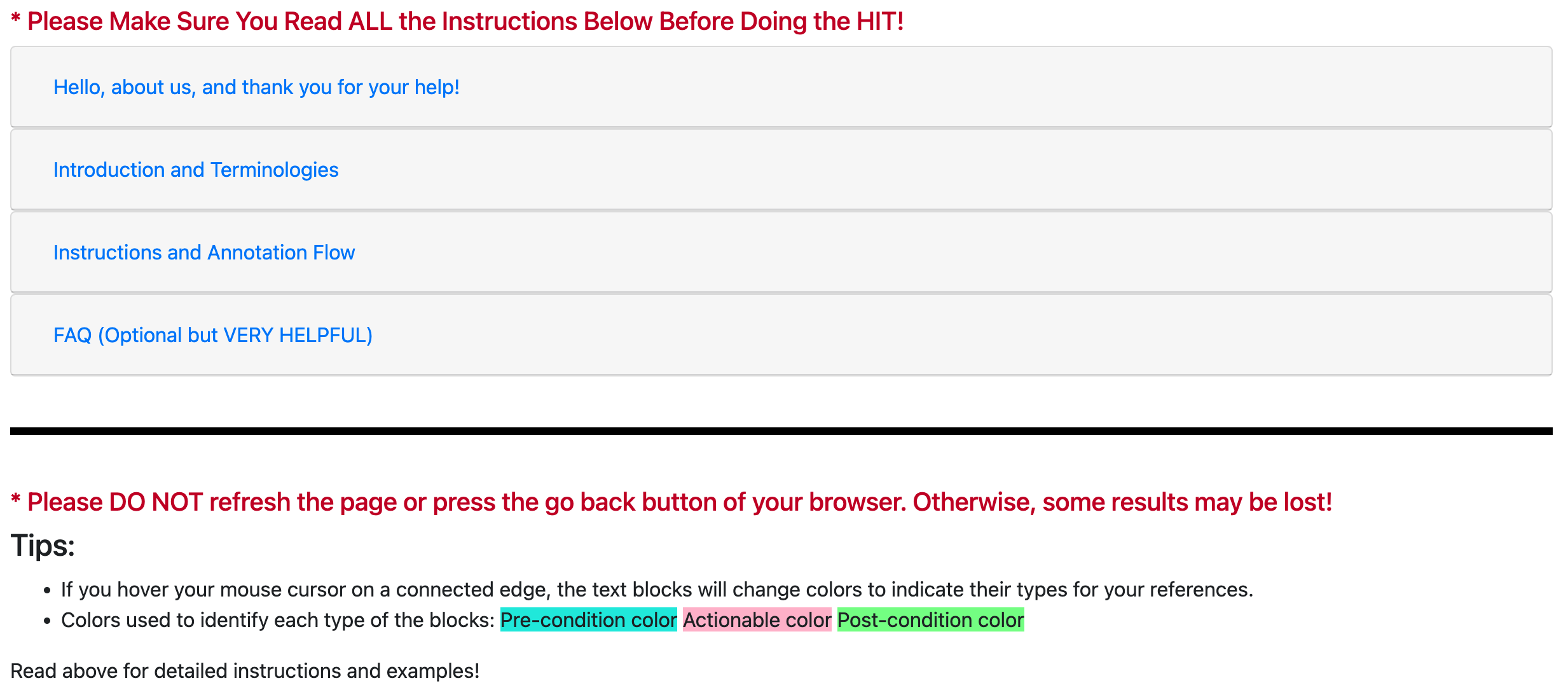}
\caption{\footnotesize Human Annotation Instruction}
\label{fig:ui_instr}
\end{subtable}

\vspace{1.em}

\begin{subtable}{\textwidth}
\centering
\centering
    \includegraphics[width=.95\columnwidth]{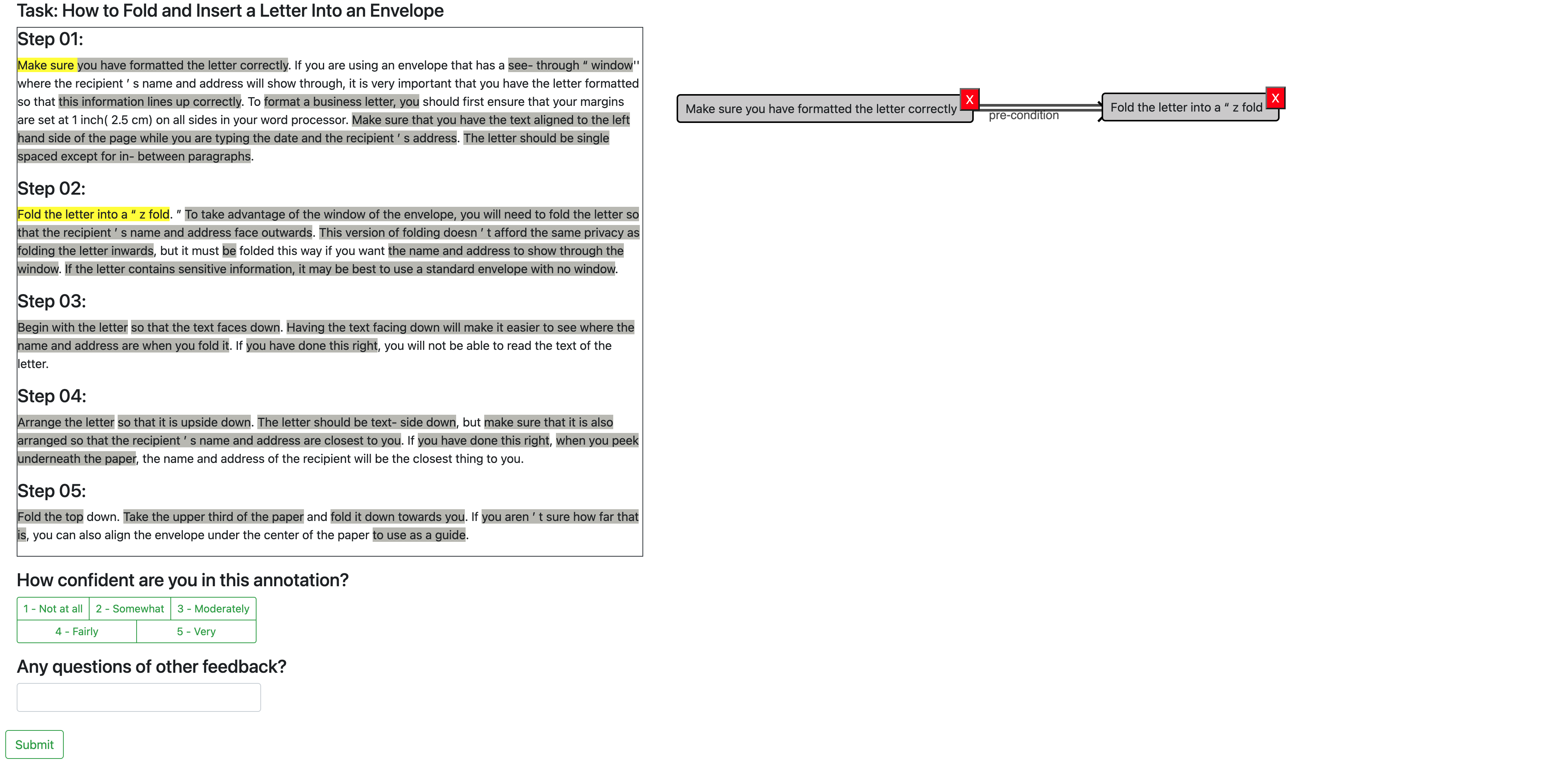}
\caption{\footnotesize Sample Annotation Interface}
\label{fig:ui_ui}
\end{subtable}

\vspace{.5em}

\caption{ \footnotesize
    \textbf{MTurk Annotation User Interface:}
    \textbf{(a)} We ask workers to follow the indicated instruction. All the blue-colored text bars on the top of the page are expandable. Workers can click to expand them for detailed instructions of the annotation task.
    \textbf{(b)} The annotation task is designed for an intuitive \textit{click/select-then-link} usage, followed by a few additional questions such as confidence level and feedback
    (this example is obtained from WikiHow dataset).
    The grey-color-highlighted text segments are postulated by the SRL, where the color of a segment will turn yellow if either being selected or cursor highlighted.
    Notice that for better illustration, the directions of the links in our paper are opposite to those in the annotation process.
}
\label{fig:ui}
\end{figure*}

\end{document}